\documentclass[sigconf,]{acmart}
\AtBeginDocument{%
  }

\setcopyright{acmlicensed}
\copyrightyear{2025}
\acmYear{2025}
\acmDOI{XXXXXXX.XXXXXXX}

\usepackage{multirow} 

\usepackage{amsfonts} 
\usepackage{amsbsy} 
\graphicspath{ {./figures/} }
\usepackage[linesnumbered,ruled,vlined]{algorithm2e} 
\usepackage{amsthm} 
\usepackage{subfig} 
\usepackage{enumitem} 
\usepackage{colortbl} 
\usepackage{siunitx} 

\usepackage{xr}
\makeatletter

\newcommand*{\addFileDependency}[1]{
\typeout{(#1)}
\@addtofilelist{#1}
\IfFileExists{#1}{}{\typeout{No file #1.}}
}\makeatother

\newcommand*{\myexternaldocument}[1]{%
\externaldocument{#1}%
\addFileDependency{#1.tex}%
\addFileDependency{#1.aux}%
}

\myexternaldocument{supplementary_material}

\newenvironment{questions}{
  \begin{enumerate}[label=\textbf{RQ\arabic*:}, leftmargin=*]
}{
  \end{enumerate}
}

\begin{document}

\title{A GAN Approach for Node Embedding in Heterogeneous Graphs Using Subgraph Sampling}


\author{Hung-Chun, Hsu}
\email{r10946017@@citi.sinica.edu.tw}
\affiliation{%
  \institution{Academia Sinica}
  \city{Nangang}
  \state{Taipei}
  \country{Taiwan}
  \postcode{11529}
}

\author{Bo-Jun, Wu}
\email{108308057@nccu.edu.tw}
\affiliation{%
  \institution{National Chenchi University}
  \city{Wenshan}
  \state{Taipei}
  \country{Taiwan}
  \postcode{116302}
}

\author{Ming-Yi, Hong}
\email{d09948005@ntu.edu.tw}
\affiliation{%
  \institution{National Taiwan University and Academia Sinica}
  \city{Da'an}
  \state{Taipei}
  \country{Taiwan}
  \postcode{10617}
}

\author{Che, Lin}
\email{chelin@ntu.edu.tw}
\affiliation{%
  \institution{National Taiwan University}
  \city{Da'an}
  \state{Taipei}
  \country{Taiwan}
  \postcode{10617}
}

\author{Chih-Yu, Wang}
\email{cywang@citi.sinica.edu.tw}
\affiliation{%
  \institution{Academia Sinica}
  \city{Nangang}
  \state{Taipei}
  \country{Taiwan}
  \postcode{11529}
}

\begin{abstract}
Graph neural networks (GNNs) face significant challenges with class imbalance, leading to biased inference results. To address this issue in heterogeneous graphs, we propose a novel framework that combines Graph Neural Network (GNN) and Generative Adversarial Network (GAN) to enhance classification for underrepresented node classes. The framework incorporates an advanced edge generation and selection module, enabling the simultaneous creation of synthetic nodes and edges through adversarial learning. Unlike previous methods, which predominantly focus on homogeneous graphs due to the difficulty of representing heterogeneous graph structures in matrix form, this approach is specifically designed for heterogeneous data. Existing solutions often rely on pre-trained models to incorporate synthetic nodes, which can lead to optimization inconsistencies and mismatches in data representation. Our framework avoids these pitfalls by generating data that aligns closely with the inherent graph topology and attributes, ensuring a more cohesive integration. Evaluations on multiple real-world datasets demonstrate the method’s superiority over baseline models, particularly in tasks focused on identifying minority node classes, with notable improvements in performance metrics such as F-score and AUC-PRC score. These findings highlight the potential of this approach for addressing key challenges in the field.
\end{abstract}




\maketitle

\section{Introduction}

Graph neural networks (GNNs) \cite{GNN_4700287} are a class of neural networks specifically designed to process graph data. GNNs show remarkable adaptability in managing highly interconnected data of varying scales, making them suitable for a wide range of domains and problems. Graphs are typically categorized as homogeneous or heterogeneous, based on the diversity of their nodes and edges. Extensive research has been conducted on both, with notable examples including the Graph Convolutional Network (GCN) \cite{DBLP:journals/corr/KipfW16} and the Graph Attention Network (GAT) \cite{veličković2018graph} for homogeneous graphs. Conversely, models like the Heterogeneous Graph Neural Network (HetGNN) \cite{10.1145/3292500.3330961} and the Heterogeneous Graph Transformer (HGT) \cite{hu2020heterogeneous} are specifically developed to address the complexities inherent in heterogeneous graphs \cite{HAN, MAGNN}.

A prominent research area in Graph Neural Networks (GNNs) focuses on addressing the node class imbalance problem \cite{DR-GCN, DPGNN, zhao2021graphsmote, TAM, mGNN, HyperIMBA}, which refers to the situation where the nodes belonging to a particular class constitutes a significant proportion of the total nodes within the graph data. Commonly, oversampling methods \cite{chawla2002smote, Borderline-SMOTE, LoRAS} are employed, using specific data sampling techniques to achieve a balanced dataset. These methods can be broadly categorized into (1) Non-generative oversampling methods, including notable examples like GraphSMOTE \cite{zhao2021graphsmote}, manipulate existing data points through transformations or interpolations, and (2) generative oversampling methods, such as ImGAGN \cite{Qu2021ImGAGN}, create new data points by sampling from distributions learned from existing data. 

Research efforts have explored integrating node attributes and topological information into generative models for graphs \cite{GraphGAN}. However, existing methods are often constrained by architectures designed for attribute-less or homogeneous graph data in matrix forms \cite{li2023graphsha}. Some strategies utilize GNN-based encoders to generate node embeddings by leveraging attributes and topology \cite{zhao2021graphsmote}, but these methods struggle to create synthetic nodes that seamlessly connect with the original graph dataset, frequently requiring additional edge predictors for establishing necessary connections. On the other hand, some approaches handle data augmentation tasks by using non-graph classifiers like MLPs to bypass the need for edge generation. However, this approach sacrifices crucial edge relations within the dataset and can lead to suboptimal task performance due to the loss of topological context. Additionally, some techniques combine generated and original data using improvised methods, typically involving fixed pre-trained models and will lead to inconsistencies and suboptimal outcomes when addressing new tasks. One strategy simplifies the connection problem by linking artificial nodes to all nodes in the original graph \cite{Qu2021ImGAGN}, allowing the GNN to learn edge weights and identify significant connections during training. However, this method increases the size of augmented graphs, resulting in a sharp rise in computational costs. Another method involves training an edge predictor to determine which nodes in the original graph should connect to synthetic nodes \cite{zhao2021graphsmote}. The challenge here is the introduction of a hyperparameter known as the edge threshold, which dictates the connections between generated and other nodes, thereby making the fine-tuning process more time-consuming. These challenges highlight the necessity for a novel, integrated approach that directly incorporates both node and edge data into the generative model’s training process.

We observe that all existing approach that incorporating synthetic nodes into the original graph to mitigate class imbalance actually face a fundamental yet critical challenge: edge imbalance. This issue is particularly pronounced in sparse graphs, where unconnected node pairs (negative edges) vastly outnumber connected pairs (positive edges). Such disparities exacerbate class imbalance during training, often leading to a significant overrepresentation of negative edges, which complicates the effective training of edge predictors for synthetic nodes and usually lead to low-quality synthetic edge construction. 

Motivated by these observations, we introduce the \textbf{FlashGAN} framework (\textbf{F}ramework of \textbf{L}ocalized Node \textbf{A}ugmentation via \textbf{S}emi-supervised Learning in \textbf{H}eterogeneous Graphs with \textbf{G}enerative \textbf{A}dversarial \textbf{N}etworks). FlashGAN is a semi-supervised data augmentation framework that learns to generate synthetic data by leveraging both labeled and unlabeled data within graphs. Additionally, FlashGAN features a novel synthetic edge filter designed for simultaneous adversarial learning of both node and edge information. It also employs a subgraph sampling mechanism \cite{sage, Cluster-GCN, GraphSAINT}, enabling it to train on graph datasets of any scale. At the same time, the use of subgraph sampling mitigates the limitations of traditional edge predictors, effectively addressing the edge imbalance problem. FlashGAN offers several significant advantages (1) It integrates node and edge data directly into the generator's training process, enhancing the GAN model’s ability to capture both structural and attribute-based features. (2) By leveraging local augmentation, FlashGAN tackles edge imbalance by embedding synthetic nodes within subgraphs. Our experiments demonstrate that models trained with edge sampling within local subgraphs outperform those trained with edge sampling across the entire graph in downstream node classification tasks. (3) The synthetic edge filter simplifies the process of optimizing edge threshold hyperparameters during the data augmentation phase, ensuring the quality of edges generated alongside nodes. In summary, our main contributions are as follows:
\begin{itemize}[noitemsep, topsep=0pt]
    \item 
    We propose a general framework for performing node augmentation with edge awareness in a heterogeneous graph, enabling simultaneous training with both node and edge information.
    \item 
    To the best of our knowledge, we are the first to design a local node augmentation and subgraph sampling method specifically to address the edge imbalance issue. This approach ensures that both node and edge imbalances are effectively managed throughout the augmentation process. 
    \item
    We enhance the GAN framework for graph augmentation tasks by incorporating a synthetic edge filter and modifying the loss function, which streamlines hyperparameter usage and ensures the quality of edges generated alongside nodes.
\end{itemize}

\section{Related Works}
The class imbalance problem in structured data has been extensively studied, with various approaches such as weight-adjusting methods \cite{7780949, focal_loss, cui2019classbalancedlossbasedeffective} and up-sampling strategies developed based on SMOTE \cite{chawla2002smote, 10.1007/978-3-540-39804-2_12, Borderline-SMOTE, 4633969, 5403368}. However, these methods often overlook the relational information inherent in graph datasets, leading to suboptimal performance compared to graph-based approaches. Several significant contributions have been made to address the class imbalance problem in graph data. GraphSMOTE \cite{zhao2021graphsmote} combines Graph Neural Networks (GNN) with interpolation to generate synthetic node embeddings. However, it relies on an additional edge predictor to integrate synthetic nodes into the graph, decoupling edge creation from the objective of generating realistic nodes and suffering from edge imbalance when using an entire graph edge sampling strategy. PC-GNN \cite{10.1145/3442381.3449989} utilizes a label-balanced sampler to enhance fraud detection by learning structural information and interactions in the graph. DR-GCN \cite{DR-GCN} addresses multi-class imbalances in graphs by implementing dual regularization and class-conditioned adversarial training. GNN-INCM \cite{huang2022graph} includes embedding clustering and graph reconstruction modules for improved representation and classification. ImGAGN \cite{Qu2021ImGAGN} employs a generator to simulate minority class attributes and topological structures; however, the assumption that synthetic nodes are connected to all original nodes necessitates the use of a massive matrix to learn the relevance of these nodes. GraphENS \cite{park2022graphens} addresses the class imbalance by synthesizing ego networks for minor class nodes and integrating a saliency-based node mixing technique. GraphSHA \cite{li2023graphsha} addresses class imbalance by enlarging the decision boundaries of minor classes by synthesizing harder minor samples. However, these methods are constrained by their focus on either homogeneous graph data or rule-based approaches for generating minority nodes. They often fail to leverage the full potential of generative models in learning complex data distributions, particularly within heterogeneous graphs. To address these gaps, we propose FlashGAN, a novel framework that overcomes these limitations and introduces a more flexible approach to adversarial learning in heterogeneous graph data. FlashGAN’s ability to integrate both node and edge information within a unified adversarial framework represents a significant advancement in tackling the class imbalance problem.

\begin{figure*}
    \includegraphics[width=0.94\textwidth]{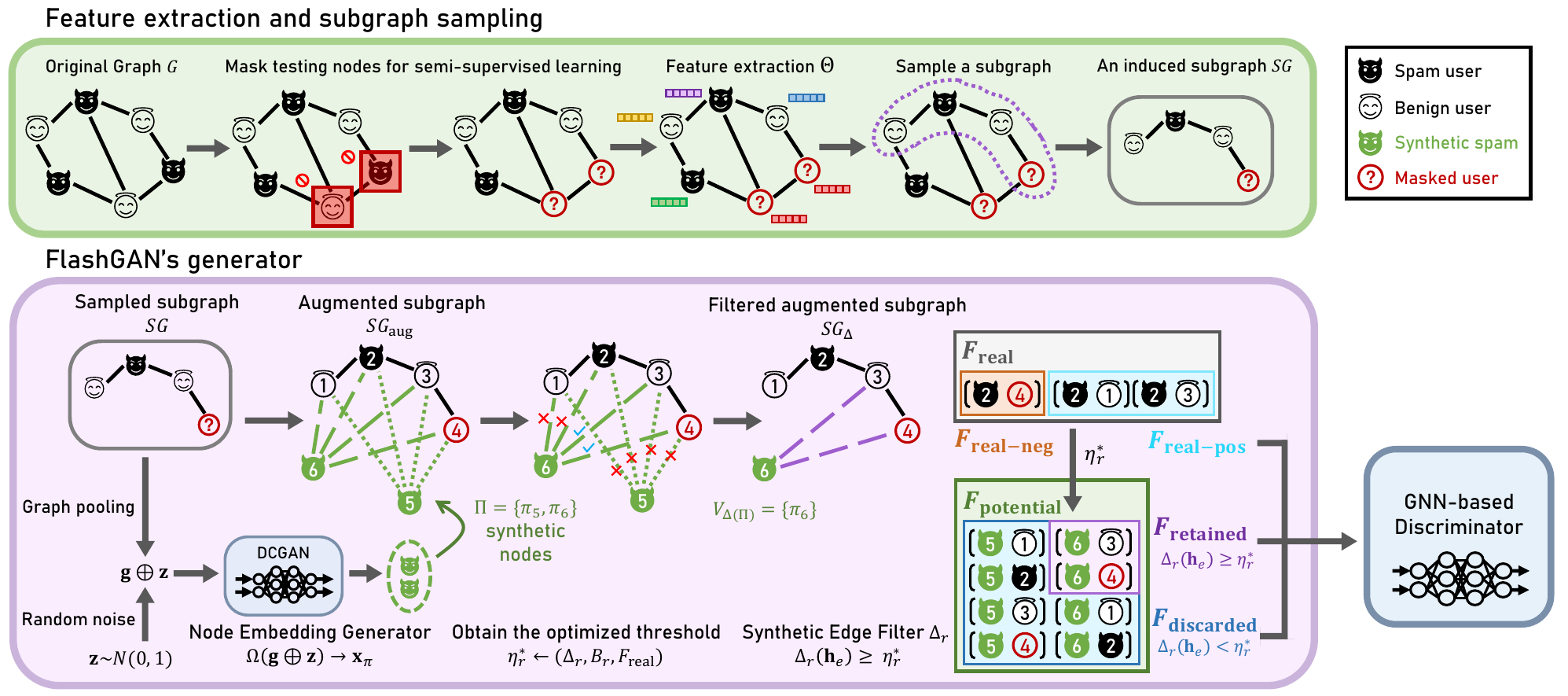}
    \caption{FlashGAN Training Workflow: FlashGAN processes subgraphs in batches. The workflow for a single subgraph is as follows: (1) Extract subgraph $SG$ from the original graph $G$. (2) Input subgraph embedding $\textbf{g}$ into the synthetic node generator $\Omega$. (3) Connect generated synthetic nodes $\Pi = \{ \pi_1, \pi_2 \}$ to all nodes in the  sampled subgraph $SG$, forming the augmented subgraph $SG_{\text{aug}}$. (4) Pass real edges $F_{\text{real}}$ through a synthetic edge filter $\Delta_r$, obtain the edge threshold $\eta_r^{*}$, and split synthetic edges $F_{\text{potential}}$ into retained and discarded groups, $F_{\text{retained}}$ and $F_{\text{discarded}}$. (5) Obtain filtered augmented subgraph $SG_{\Delta}$ with selected synthetic nodes and retained edges. (6) Update the generator based on the discriminator’s classification of retained and discarded edges.}
\label{FlashGAN_overview}
\end{figure*}

\section{Preliminaries and Definition}

\subsection{Imbalance Ratio and Imbalanced Graph}

We define a heterogeneous attribute graph as $G = (V, E, \mathbf{X}_V)$, with $\phi\left( \nu \right)\mathcal{\,:\, } V \rightarrow \mathcal{A}$ and $\psi\left( e \right)\mathcal{\,:\,} E \rightarrow \mathcal{R}$ as the node and edge type mapping functions, respectively. The node attribute matrix $\mathbf{X}_V \in \mathbb{R}^{n \times d}$, where $n$ is the number of nodes and $d$ is the attribute dimension. Node class labels are denoted by $\mathbf{Y} \in \mathbb{R}^n$. For a graph $G$ with $m$ classes, the set of node classes is $\{C_1, \dots, C_m\}$. Let $\lvert C_i \rvert$ denote the size of the $i$-th class. The smallest and largest classes are defined as $C_\text{min}$ and $C_\text{maj}$, respectively. The imbalance ratio $\text{IR} = \frac{\lvert C_\text{min} \rvert}{\lvert C_\text{maj} \rvert}$ measures the degree of class imbalance, with a graph $G$ considered imbalanced if $\lvert C_\text{min} \rvert$ is significantly smaller than $\lvert C_\text{maj} \rvert$.

\subsection{Generative Adversarial Network}
Generative Adversarial Networks (GANs) \cite{goodfellow2014generative} are powerful generative models widely used in data augmentation \cite{radford2015unsupervised, mirza2014conditional, DCGAN}. A typical GAN framework consists of a generator $G$ and a discriminator $D$, which compete in a min-max game to replicate the underlying data distribution. The objective function is:
\begin{equation}
\min \limits_{G} \max \limits_{D}V(D, G) = \mathbb{E} \Bigl[ \text{logD}(\mathbf{x}) \Bigr] + \mathbb{E}\Bigl [\text{log}\Bigl(1-\text{D}\bigl(\text{G}(\mathbf{z}) \bigr) \Bigr) \Bigr]
\end{equation}
Through iterative training, $G$ aims to generate samples that $D$ cannot distinguish from real data, ultimately making the generated samples indistinguishable from actual data.

\subsection{Node Classification on Imbalanced Graph}

The node classification task in Graph Neural Networks (GNNs) predicts node labels within a given graph. In the case of imbalanced node classes, where $\lvert C_\text{maj}\rvert \gg \lvert C_\text{min}\rvert$, the task is to learn a classifier $\mathcal{F}_{\lambda}(V \mid G) \rightarrow \mathbf{Y}$ that accurately predicts labels for both majority and minority classes. The classifier’s objective function is:
\begin{equation}
    \mathop{\arg\max}\limits_{\lambda} f \biggl[~ \Bigl ( \mathcal{F}_{\lambda}(a \mid G), y_a \Bigr ), \Bigl ( \mathcal{F}_{\lambda}(b \mid G), y_b \Bigr ), \forall a \in A, \forall b \in B ~ \biggr]
\end{equation} where $A = \{v \in V \mid y_v = C_{\text{min}} \}$, $B = \{u \in V \mid y_u = C_{\text{maj}} \}$, $y_u$ is the class label of node $u$, and $f$ represents any performance metric, such as F-score.

\section{Methodology}

In this section, we present the architecture of FlashGAN, designed to streamline node generation, simplify edge management, and enhance graph augmentation with a focus on minority node connections. FlashGAN operates by embedding synthetic nodes into local subgraphs, determining node placement and features based on the subgraph's context. For example, in a social network dataset aimed at spam detection \cite{mcauley2013amateurs, GraphConsis, GEM}, synthetic spam user data is generated and integrated into the training set, balancing the distribution between benign and spam users for improved model performance. The architecture of FlashGAN follows the classical generative adversarial network structure, consisting of a generator and a discriminator. The generator includes a node embedding generator, which produces synthetic nodes for connection within the sampled subgraph, and a synthetic edge filter, which selectively retains or discards synthetic edges based on their potential to deceive the discriminator. The discriminator then distinguishes between real edges in the original subgraph and synthetic edges formed by the insertion of synthetic nodes. As illustrated in Figure \ref{FlashGAN_overview}, we use a toy example of a user-user network $G$ with a single edge type $r=\text{'user-user'}$ to clarify each component of FlashGAN. This setup mirrors real-world spam detection scenarios, where spam users, typically minority nodes, are distinguished from benign users, who represent the majority class.

\subsection{Feature Extractor}
To prevent the raw feature space of nodes from being too sparse and to avoid the challenge of a low dimensionality, which can hinder the generator of FlashGAN from learning the true distribution of the data, we employ a feature extractor to obtain node representations that better reflect node properties and graph topology \cite{zhao2021graphsmote}. Let $\Theta$ be a general Heterogeneous Graph Neural Network (HetGNN) model. For a heterogeneous graph $G = (V, E, \mathbf{X}_V)$ with node feature matrix $\mathbf{X}_V$, we could obtain its node hidden representation matrix $\text{H}_{V}$ by:
\begin{equation}
    G \Bigl(V, E, \mathbf{H}_V \Bigr) = \Theta \Bigl(G \bigl(V, E, \mathbf{X}_V \bigr) \Bigr)  \label{eq_feature_extract}
\end{equation}
where $\mathbf{H}_{V} \in  \mathbb{R}^{\rvert V \lvert \times b}$, $\rvert V \lvert$ is the size of the node set $V$ and $b$ is the size of the hidden representation of nodes.

Any HetGNN model can serve as the backbone for feature extraction in FlashGAN. In this study, we employ the Heterogeneous Graph Transformer (HGT) \cite{hu2020heterogeneous} to learn node representations from the training and validation nodes, ensuring that testing node labels are removed to prevent data leakage. FlashGAN is then trained on these new node representations. After training, FlashGAN’s generator augments the graph by adding synthetic nodes to the training portion. Finally, during the downstream testing phase, we reintegrate the testing node labels into the augmented graph and train a node classification model to evaluate the performance on this enhanced dataset.

\subsection{Node Embedding Generator}
Before generating synthetic nodes to balance the node classes in the original dataset, we perform subgraph sampling from the original graph (\textit{e.g.}, one-hop subgraph sampling) \cite{park2022graphens, li2023graphsha}. During model training, these sampled subgraphs are fed into the model in batches. For a sampled subgraph $SG = (V_{SG}, E_{SG})$, the node embedding generator creates synthetic nodes and their embeddings for integration. Let $k$ synthetic nodes be generated for integration into a sampled subgraph $SG$ to accomplish the subgraph augmentation process. We introduce a learnable DCGAN generator \cite{DCGAN} $\Omega \left( \mathbf{g} \oplus \mathbf{z} \right): \mathbb{R}^{b+d} \rightarrow \mathbb{R}^{b}$ and modify it to act as a node embedding generator. To generate a synthetic node represented as $\pi$ with its corresponding node attributes $\mathbf{x}_{\pi}$, we concatenate the subgraph representation $\mathbf{g} \in \mathbb{R}^{b}$ obtained from global pooling with a $d$-dimensional random noise $\mathbf{z} \in \mathbb{R}^{d}$ sampled from a standard normal distribution $\mathcal{N}\left(0, 1\right)$ into the generator $\Omega$. The $k$ nodes inserted into the subgraph $SG$ and their respective node attributes are represented by $\Pi = \{\pi_1, \pi_2, ..., \pi_k\}$ and $\mathbf{H}_{\Pi} = \{\mathbf{h}_{\pi_1}, \mathbf{h}_{\pi_2}, ..., \mathbf{h}_{\pi_k} \}$, respectively. Finally, we let all synthetic nodes be connected to all real nodes in the original subgraph. Since every synthetic node was, by default, connected to all other nodes belonging to the corresponding subgraph. We represent these default-connected edges with the set $F_{\text{potential}}$ and refer to them as potential synthetic edges. After adding synthetic nodes to the subgraph $SG$, we refer to this subgraph as the \textit{\textbf{augmented subgraph}} $\boldsymbol{SG_{\text{aug}}}$, denoted by \eqref{eq_augmented_subgraph}:
\begin{equation}
    SG_{\text{aug}} = (V_{SG_{\text{aug}}}, E_{SG_{\text{aug}}}) = (V_{SG} \cup \Pi, E_{SG} \cup  F_{\text{potential}})
\label{eq_augmented_subgraph}
\end{equation}
where $F_{\text{potential}} = \{(v_\text{src}, v_\text{dst}), (v_\text{dst}, v_{\text{src}})~\mid~v_{\text{src}}\in V_{SG}, v_{\text{dst}} \in \Pi \}$.

\subsection{Adversarial Synthetic Edge Filter}
A synthetic node is successfully added to a subgraph due to the establishment of associated potential edges; however, not all synthetic edges are similar enough to real edges to be retained in the subgraph, so we aim to remove redundant synthetic edges if they are of poor quality. Consequently, the synthetic edge filter is specifically tasked with preserving potential synthetic edges that are sufficiently similar to real edges to deceive the discriminator. Specifically, for each edge type $r \in \mathcal{R}$ (\textit{e.g.}, 'user-user' or 'user-product') in a heterogeneous graph, we introduce a synthetic edge filter $\Delta_r$.
\begin{equation}
    \Delta_r(\mathbf{h}_{e(r)}) = \sigma(\mathbf{h}_u^\top \mathbf{W}_{\Delta_r}^{\vphantom{\top}} \mathbf{h}_v^{\vphantom{\top}}) = p
    \label{eq_syn_edge_filter}
\end{equation}
The synthetic edge filter $\Delta_r$ is modeled with learnable weighted matrix $\mathbf{W}_{\Delta_r}$ \cite{sage} to compute a probability $p \in [0, 1]$ for an edge $e(r) = (u, v)$, reflecting the likelihood of retaining the edge $e(r)$. The edge embedding $\mathbf{h}_{e(r)} = (\mathbf{h}_{u}, \mathbf{h}_{v})$, defined as the node embeddings from $\mathbf{H}_V[:, u]$ and $\mathbf{H}_V[:, v]$, is learned during the previous feature extraction step described in \eqref{eq_feature_extract}. If the likelihood $p$ falls below the threshold $\eta_r$, the edge $e(r)$ is discarded from the augmented subgraph $SG_{\text{aug}}$. Therefore, the potential edge set $F$ can be further divided into $F_{\text{retained}}$ and $F_{\text{discarded}}$ by the edge threshold value $\eta_r$. These two edge sets respectively represent the synthetic edges that are retained and those that are discarded by the edge filter. 

Next, we seek to determine the optimal threshold value $ \eta_r$. Typically, this threshold is treated as a hyperparameter, adjusted through experimentation, but this approach increases computational costs. Given that FlashGAN’s primary goal is to generate synthetic nodes and edges that closely resemble real ones, we propose determining $ \eta_r$ by using existing edges in subgraphs $SG$ that include real nodes from the minority class $C_\text{min}$. The motivation stems from the premise that if an edge filter can effectively retain (or discard) synthetic edges that are sufficiently similar (or dissimilar) to deceive (or not deceive) the discriminator, it should also precisely determine whether node pairs containing real minority nodes form positive or negative edges, \textit{i.e.}, whether the node pairs are connected by edges. Thus, the ideal threshold should differentiate real positive and negative edges precisely. The computation of the optimized threshold $\eta_r^{}$ is detailed in \eqref{optimized_eta} and \eqref{binary_step_function}, where both positive and negative edges containing real minority nodes are processed through the synthetic edge filter. The resulting probabilities are binarized using a binary step function $B_r$, with the threshold $\eta_r^{}$ that minimizes loss being selected for filtering synthetic edges.

\begin{equation}
    \label{optimized_eta}
    \eta_r^{*} = \mathop{\arg\min}\limits_{\eta_r \in [0, 1]} \Biggl\{ \texttt{loss function} \biggl[ \Bigl( B_r(\mathbf{h}_{e'} ), y(e') \Bigr), \forall e' \in F_{\text{real}} \biggr] \Biggr\}
\end{equation}

\begin{equation}
    \label{binary_step_function}
    B_r(\mathbf{h}_{e'}) \overset{\text{def}}{=} B_r(\mathbf{h}_{e'} \mid \Delta_r, \eta_r) = 
    \begin{cases} 
    1 & \text{if } \Delta_r({\mathbf{h}_{e'}}) \geq \eta_r \\
    0 & \text{if } \Delta_r({\mathbf{h}_{e'}}) < \eta_r
    \end{cases}
\end{equation}
In implementing the aforementioned loss function, we utilize binary cross-entropy loss, where $\mathbf{h}_{e'}$ represents the edge embeddings of edge $e' \in F_{\text{real}}$, which includes real minority nodes but excludes synthetic nodes. The edge sets $F_{\text{real-pos}}$ and $F_{\text{real-neg}}$, as defined in \eqref{real_edge}, denote the positive and negative edges, respectively, within subgraph $SG$, containing real nodes that authentically belong to the minority class $C_{\text{min}}$.
\begin{equation}
    \label{real_edge}
\begin{aligned}
    F_{\text{real}} &= F_{\text{real-pos}} \cup F_{\text{real-neg}} \\
    F_{\text{real-pos}} &= \left\{ (u,v) \in E_{SG} \mid u, v \in V_{SG}, ~\text{s.t.,}~y_u = \text{C}_{\text{min}} \lor y_v = \text{C}_{\text{min}} \right\} \\
    F_{\text{real-neg}} &= \left\{ (u,v) \notin E_{SG} \mid u, v \in V_{SG}, ~\text{s.t.,}~y_u = \text{C}_{\text{min}} \lor y_v = \text{C}_{\text{min}} \right\} \\
\end{aligned}
\end{equation}

As defined in \eqref{potential_edge}, the potential synthetic edge set $F_{\text{potential}}$ is divided into $F_{\text{retained}}$ and $F_{\text{discarded}}$ based on the edge threshold value $\eta_r^{*}$ derived from \eqref{optimized_eta} and \eqref{binary_step_function}. $F_{\text{retained}}$ represents the edges that are retained, while $F_{\text{discarded}}$ indicates the synthetic edges removed from the augmented subgraph $SG_{\text{aug}}$.

\begin{equation}
    \label{potential_edge}
\begin{aligned}
    F_{\text{potential}} &= F_{\text{retained}} \cup F_{\text{discarded}} \\
    F_{\text{retained}} &= \{e_r \in F_{\text{potential}} \subset E_{SG_{\text{aug}}} \mid \Delta_r(\mathbf{h}_{e(r)}) \geq \eta_r^{*} \} \\
    F_{\text{discarded}} &= \{e_r \in F_{\text{potential}} \subset E_{SG_{\text{aug}}} \mid \Delta_r(\mathbf{h}_{e(r)}) < \eta_r^{*} \}
\end{aligned}
\end{equation}

In this context, the necessity to convert edge-retaining probabilities that fall within the $[0, 1]$ range into binary values arises from the requirement to use any GNN-based discriminator for training or node classification in the downstream tasks. Specifically, probabilistic edges must be transformed into a binary representation, where connected is indicated by 1 and not connected by 0. After applying edge filter $\Delta_r$ to all the potential edges $e(r) \in F_{\text{potential}} \subset E_{SG_{\text{aug}}}$, we will eventually obtain a \textit{\textbf{filtered augmented subgraph}} $SG_{\Delta}$.
\begin{equation}
    SG_{\Delta} = (V_{SG_\Delta}, E_{SG_{\Delta}}) = (V_{SG} \cup V_{\Delta(\Pi)}, E_{SG} \cup F_{\text{retained}})
\end{equation}
where $V_{\Delta(\Pi)} = \left\{ \pi_i \in \Pi \mid \exists ~u \in V_{SG} ~\text{s.t.,}~ (u, \pi_i) \in E_{SG_{\Delta}}  \right\}$.

\subsection{GNN-based Discriminator}
To best utilize synthetic edge information, we employ a GNN-based discriminator to distinguish between real edges, which contain real spam users, and synthetic edges, which contain synthetic spam users, within each filtered augmented subgraph $SG_{\Delta}$. We utilize HGT as the backbone GNN model for the discriminator, denoted as $\Theta_{\textsc{Dis}}$, and apply a learnable weighted matrix $\mathbf{W}_{\textsc{Dis}}$ that acts as the edge classifier. The function of the discriminator is mathematically described in \eqref{discriminator_HGNN} and \eqref{discriminator_HGNN_edge_classifier}, where $\mathbf{h}^{*}_u = \mathbf{H}^{*}_{V_{SG_{\Delta}}}[:, u]$ and $\mathbf{h}^{*}_v$ $=$ \\$\mathbf{H}^{*}_{V_{SG_{\Delta}}}[:, v]$ are the node embeddings learned by $\Theta_{\textsc{Dis}}$. 

\begin{equation}
    SG_{\Delta} \Bigl(V_{SG_{\Delta}}, E_{SG_{\Delta}}, \mathbf{H}^{*}_{V_{SG_{\Delta}}} \Bigr) = \Theta_{\textsc{Dis}} \Bigl(SG_{\Delta}\bigl (V_{SG_{\Delta}}, E_{SG_{\Delta}}, \mathbf{H}_{V_{SG_{\Delta}}} \bigr) \Bigr)  \label{discriminator_HGNN}
\end{equation}
\begin{equation}
    \sigma(\mathbf{h}_u^{*\top} \mathbf{W}_{\textsc{Dis}}^{\vphantom{\top}} \mathbf{h}_v^{*\vphantom{\top}}) = q \quad \texttt{// real or fake}
    \label{discriminator_HGNN_edge_classifier}
\end{equation}

Note that the node embeddings learned by the discriminator are denoted with an asterisk (*) to avoid confusion with the hidden representation matrix $\mathbf{H}_{V}$ learned in \eqref{eq_feature_extract}.

\subsection{Downstream Node Classification Task}
After training FlashGAN, the generator $\textsc{Gen}$ can augment the original graph by adding synthetic nodes. The number of synthetic nodes to add is typically determined by the imbalance ratio in the training data. For a graph $G$ with class imbalance, let $M$ represent the number of nodes in the majority class $C_{\text{maj}}$, $m$ the number in the minority class $C_{\text{min}}$, and $m'$ the number in $C_{\text{min}}$ after augmentation. To achieve an imbalance ratio of $\text{IR} = \alpha = \frac{m'}{M}$, the required number of synthetic nodes is $m' - m = \alpha \times M - m$. This augmented graph can then be used for downstream tasks such as node classification. 

\section{Optimization of FlashGAN}
In this section, we define the objective function of FlashGAN and the corresponding loss functions for the generator and discriminator. Unlike GAN-based generative oversampling methods focusing on node-level generation and discrimination, FlashGAN stands out as an edge-aware node augmentation framework. While FlashGAN introduces additional nodes, the optimization goal of its generator is to make the edges created by these synthetic nodes resemble the real edges. On the other hand, the discriminator aims to distinguish between these edges. 
\subsection{Loss function of Generator}

For the $i$-th filtered augmented subgraph $SG_{\Delta}^{(i)}$ with its previous incarnation $SG_{\text{aug}}^{(i)}$, the loss function of generator $\textsc{Gen}$ can be expressed by \eqref{eq_generator_loss_function}, 
\begin{equation}
\begin{aligned}
    \mathcal{L}^{(i)}_{\textsc{Gen}} =&-\dfrac{1}{\lvert F_{\text{real}}^{(i)} \rvert} \biggl\{ \sum\limits_{e \in F_{\text{real-pos}}^{(i)} } \text{log} \bigl[ \Delta(\mathbf{h}_e) \bigr] + \sum\limits_{e \in F_{\text{real-neg}}^{(i)} } \text{log} \bigl[ 1-\Delta(\mathbf{h}_e) \bigr] \biggr\} \\
    &-\dfrac{1}{\lvert F_{\text{retained}}^{(i)} \rvert}\sum\limits_{e \in F_{\text{retained}}^{(i)} } \text{log} \bigl[ \textsc{Dis}(\mathbf{h}_e) \bigr] \\
    &-\dfrac{1}{\lvert F_{\text{discarded}}^{(i)} \rvert}\sum\limits_{e \in F_{\text{discarded}}^{(i)} } \text{log} \bigl[ 1 - \textsc{Dis}(\mathbf{h}_e) \bigr]
    \label{eq_generator_loss_function}
\end{aligned}
\end{equation}

The three-term generator loss function of FlashGAN is intricately designed to facilitate the generation of high-quality synthetic nodes and edges. As detailed in \eqref{eq_generator_loss_function}, the loss function is bifurcated into two principal components: the first term and a combined second and third term. Each component of the loss function serves a distinct role within the FlashGAN framework. Specifically, the first term compels the synthetic edge filter to discern real edges, allowing us to subsequently determine an optimal threshold $\eta_r^{*}$ to segregate potential synthetic edges in $F_{\text{potential}}$. In contrast, the second and third terms equip the synthetic edge filter with the ability to identify and retain synthetic edges that could deceive the discriminator. Moreover, when the edge filter erroneously retains synthetic edges that fail to deceive the discriminator, the second term imposes a significant penalty. Conversely, if the filter erroneously discards synthetic edges that could have deceived the discriminator, the third term triggers a substantial penalty. Through this mechanism, the synthetic edge filter progressively learns to maintain synthetic edges that can mislead the discriminator, ensuring that only edges analogous to real ones are preserved. This approach advances beyond traditional methods that limit adversarial training to node attribute levels by integrating synthetic edge construction directly into the GAN model's optimization process, thus enabling the simultaneous generation of synthetic nodes and edges.

\subsection{Loss function of Discriminator}
As mentioned earlier, the goal of the discriminator is to distinguish between real and synthetic edges. The synthetic edges refer to the set of edges $F_{\text{retained}}^{(i)}$, which corresponds to the subgraph $SG^{(i)}$ that has successfully embedded synthetic nodes $\pi_i$. On the other hand, the real edges consist of the set of edges within the subgraph $SG^{(i)}$ that originally existed, including edges involving minority class nodes, denoted by $F_{\text{real-pos}}^{(i)}$. The loss function of discriminator $\textsc{Dis}$ can be expressed by \eqref{eq_discriminator_loss_function}, 
\begin{equation}
\begin{aligned}
    \mathcal{L}_{\textsc{Dis}}^{(i)} =&- \dfrac{1}{\lvert F_{\text{real-pos}}^{(i)} \rvert}\sum\limits_{e\in F_{\text{real-pos}}^{(i)}} \text{log} \bigl[ \textsc{Dis}(\mathbf{h}_e) \bigr] \\
    &-\dfrac{1}{\lvert F_{\text{retained}}^{(i)} \rvert}\sum\limits_{e \in F_{\text{retained}}^{(i)} } \text{log} \bigl[ 1 - \textsc{Dis}(\mathbf{h}_e) \bigr] \label{eq_discriminator_loss_function}
\end{aligned}
\end{equation}

\subsection{Objective Function of FlashGAN}
The objective function of FlashGAN can be represented by \eqref{obj_function_of_FlashGAN}, where $\mathbf{h}_x$ is the edge embedding of the edge $x$ belonging to the corresponding filtered augmented subgraph $SG_{\Delta}^{(i)}$. $\mathcal{P}$ and $\mathcal{Q}_{\textsc{Gen}}$ are the edge embedding distributions of the edge sets induced by real minority nodes and synthetic minority nodes, respectively represented by $\mathcal{P} = \cup_{i} F_{\text{real-pos}}^{(i)}$ and $\mathcal{Q}_{\textsc{Gen}} = \cup_{j} F_{\text{retained}}^{(j)}$. The subscript symbol $\textsc{Gen}$ represents that the formation is affected by the generator $\textsc{Gen}(\Omega, \Delta)$.
\begin{equation}
\begin{aligned}
    \min \limits_{\textsc{Gen}} \max \limits_{\textsc{Dis}}V \Bigl( \textsc{Dis}, \textsc{Gen} \Bigr) &=\mathbb{E}_{p\sim \mathcal{P}} \Bigl[ \text{log}\textsc{Dis}(\mathbf{h}_p) \Bigr] \\
    &+ \mathbb{E}_{q \sim \mathcal{Q}_{\textsc{Gen}}} \Bigl[ \text{log} \Bigl( 1-\textsc{Dis}(\mathbf{h}_q) \Bigr) \Bigr] \label{obj_function_of_FlashGAN}
\end{aligned}
\end{equation}

\subsection{Training Procedure}
Due to space limitations, the pseudocode for FlashGAN is included in the supplementary materials, with the training pipeline outlined in Algorithm 1. The training process of FlashGAN begins by extracting preliminary node representations via a feature extractor. Sampled subgraphs are then augmented with synthetic nodes and edges, as detailed in Algorithm 2. These augmented subgraphs are then fed into the generator and discriminator, where parameters are iteratively updated using their respective loss functions.

\section{Experiments}
In this section, we conduct extensive experiments to evaluate FlashGAN's performance in class-imbalanced node classification, addressing the following research questions:

\begin{questions}
    \item How does FlashGAN’s performance compare to that of conventional methods and GNN baselines in various metrics?
    \item How does varying the imbalance ratio, specifically the number of synthetic nodes added to the augmented graph, impact minority node classification accuracy?
    \item Does the synthetic edge filter reliably select synthetic edges that optimize node classification performance?
    \item What is the effectiveness of the synthetic nodes and edges generated by FlashGAN on overall node classification, and how significantly do they enhance performance?
\end{questions}

\begin{table*}
\centering
\caption{Node classification results ($\pm$std) on Amazon and Yelp for 10 runs. Top results are in bold and second-best are \underline{underlined}.}\label{tb_result}
\begin{tabular}{@{}l|cccc|cccc@{}} 
\toprule
\toprule
& \multicolumn{4}{c|}{Amazon} & \multicolumn{4}{c}{Yelp} \\
\cmidrule(lr){2-5} \cmidrule(lr){6-9}
Method &  AUC-PRC & AUC-ROC & F-Score & Accuracy & AUC-PRC & AUC-ROC & F-Score & Accuracy\\
\midrule
Original & 41.51 $\scriptstyle{\pm0.048}$ & 83.49 $\scriptstyle{\pm0.020}$ & 37.70 $\scriptstyle{\pm0.044}$ & 92.28 $\scriptstyle{\pm0.010}$ & 37.15 $\scriptstyle{\pm0.024}$ & 74.53 $\scriptstyle{\pm0.014}$ & 14.19 $\scriptstyle{\pm0.086}$ & 78.56 $\scriptstyle{\pm0.006}$\\
\midrule
Reweight & 38.92 $\scriptstyle{\pm0.025}$ & 84.07 $\scriptstyle{\pm0.018}$ & \underline{41.93} $\scriptstyle{\pm0.033}$ & 91.54 $\scriptstyle{\pm0.007}$ & 31.17 $\scriptstyle{\pm0.069}$ & 65.47 $\scriptstyle{\pm0.121}$ & 22.51 $\scriptstyle{\pm0.154}$ & 76.69 $\scriptstyle{\pm0.016}$ \\
Focal Loss & 31.00 $\scriptstyle{\pm0.035}$ & 66.56 $\scriptstyle{\pm0.081}$ & 38.19 $\scriptstyle{\pm0.063}$ & 92.02 $\scriptstyle{\pm0.005}$ & 25.09 $\scriptstyle{\pm0.077}$ & 53.11 $\scriptstyle{\pm0.129}$ & 15.43 $\scriptstyle{\pm0.104}$ & 78.43 $\scriptstyle{\pm0.014}$\\
\midrule
Oversampling & 42.08 $\scriptstyle{\pm0.048}$ & \underline{86.37} $\scriptstyle{\pm0.011}$ & 41.34 $\scriptstyle{\pm0.044}$ & 92.05 $\scriptstyle{\pm0.004}$ & 38.69 $\scriptstyle{\pm0.038}$ & 74.81 $\scriptstyle{\pm0.022}$ & 22.56 $\scriptstyle{\pm0.083}$ & 78.39 $\scriptstyle{\pm0.013}$ \\
SMOTE & 41.06 $\scriptstyle{\pm0.021}$ & 86.26 $\scriptstyle{\pm0.012}$ & 40.05 $\scriptstyle{\pm0.040}$ & \underline{92.48} $\scriptstyle{\pm0.005}$ & 37.28 $\scriptstyle{\pm0.025}$ & 74.45 $\scriptstyle{\pm0.016}$ & 19.13 $\scriptstyle{\pm0.088}$ & 78.08 $\scriptstyle{\pm0.007}$ \\
\midrule
PC-GNN & 39.90 $\scriptstyle{\pm0.042}$ & 85.80 $\scriptstyle{\pm0.010}$ & 37.48 $\scriptstyle{\pm0.106}$ & 90.07 $\scriptstyle{\pm0.023}$ & 38.37 $\scriptstyle{\pm0.011}$ & \textbf{76.23} $\scriptstyle{\pm0.007}$ & 9.71 $\scriptstyle{\pm0.105}$ & \textbf{78.95} $\scriptstyle{\pm0.003}$ \\
ImGAGN & 15.56 $\scriptstyle{\pm0.012}$ & 56.83 $\scriptstyle{\pm0.006}$ & 16.95 $\scriptstyle{\pm0.050}$ & 89.16 $\scriptstyle{\pm0.011}$ & 21.87 $\scriptstyle{\pm0.004}$ & 52.92 $\scriptstyle{\pm0.007}$ & 20.53 $\scriptstyle{\pm0.147}$ & 50.02 $\scriptstyle{\pm0.276}$ \\
GraphSMOTE & \underline{44.62} $\scriptstyle{\pm0.023}$ & 86.05 $\scriptstyle{\pm0.014}$ & 40.80 $\scriptstyle{\pm0.032}$ & 92.48 $\scriptstyle{\pm0.005}$ & 37.08 $\scriptstyle{\pm0.489}$ & 74.49 $\scriptstyle{\pm0.326}$ & 13.83 $\scriptstyle{\pm5.168}$ & 78.45 $\scriptstyle{\pm0.427}$ \\
GraphENS & 43.61 $\scriptstyle{\pm0.028}$ & 65.60 $\scriptstyle{\pm0.040}$ & 25.13 $\scriptstyle{\pm0.063}$ & 65.60 $\scriptstyle{\pm0.040}$ & \underline{39.40} $\scriptstyle{\pm0.072}$ & 74.52 $\scriptstyle{\pm0.020}$ & 19.54 $\scriptstyle{\pm0.079}$ & 74.52 $\scriptstyle{\pm0.020}$\\
GraphSHA & 34.87 $\scriptstyle{\pm5.60}$ & 65.02 $\scriptstyle{\pm4.11}$ & 29.02 $\scriptstyle{\pm4.12}$ & 83.80 $\scriptstyle{\pm2.05}$ & 36.18 $\scriptstyle{\pm12.34}$ & 56.44 $\scriptstyle{\pm6.17}$ & \underline{24.76} $\scriptstyle{\pm16.35}$ & 72.64 $\scriptstyle{\pm4.45}$ \\
\midrule

\textbf{FlashGAN}$\text{\scriptsize  -R.Walk}$ & 46.07 $\scriptstyle{\pm0.024}$ & 87.98 $\scriptstyle{\pm0.006}$ & 46.71 $\scriptstyle{\pm0.018}$ & 91.99 $\scriptstyle{\pm0.007}$ & 37.68 $\scriptstyle{\pm0.012}$ & 74.67 $\scriptstyle{\pm0.009}$ & 18.58 $\scriptstyle{\pm0.089}$ & \underline{78.75} $\scriptstyle{\pm0.006}$ \\

\textbf{FlashGAN}$\text{\scriptsize -LPnE}$ & 47.81 $\scriptstyle{\pm0.034}$ & 88.06 $\scriptstyle{\pm0.003}$ & 44.34 $\scriptstyle{\pm0.037}$ & 92.26 $\scriptstyle{\pm0.005}$ & 37.35 $\scriptstyle{\pm0.026}$ & 74.52 $\scriptstyle{\pm0.008}$ & 17.43 $\scriptstyle{\pm0.059}$ & 78.06 $\scriptstyle{\pm0.011}$
 \\
\textbf{FlashGAN}$\text{\scriptsize -RSSE}$  & 48.73 $\scriptstyle{\pm0.018}$ & 88.10 $\scriptstyle{\pm0.003}$ & 46.17 $\scriptstyle{\pm0.013}$ & 91.88 $\scriptstyle{\pm0.004}$ & 37.21 $\scriptstyle{\pm0.025}$ & 74.39 $\scriptstyle{\pm0.001}$ & 18.64 $\scriptstyle{\pm0.010}$ & 78.60 $\scriptstyle{\pm0.007}$
 \\ 
\midrule

\cellcolor{gray!30} \textbf{FlashGAN} & \cellcolor{gray!30} \textbf{52.31} $\scriptstyle{\pm0.013}$ & \cellcolor{gray!30} \textbf{88.94} $\scriptstyle{\pm0.006}$ & \cellcolor{gray!30} \textbf{51.40} $\scriptstyle{\pm0.019}$ & \cellcolor{gray!30} \textbf{92.79} $\scriptstyle{\pm0.004}$ & \cellcolor{gray!30} \textbf{40.41} $\scriptstyle{\pm0.013}$ & \cellcolor{gray!30} \underline{74.93} $\scriptstyle{\pm0.009}$ & \cellcolor{gray!30} \textbf{32.05} $\scriptstyle{\pm0.037}$ & \cellcolor{gray!30} \underline{78.75} $\scriptstyle{\pm0.011}$ \\ 
\bottomrule
\bottomrule
\end{tabular}
\end{table*}

\subsection{Experimental Setup}
\subsubsection{Datasets}
We employ FlashGAN to address the class imbalance problem in real-world datasets, Amazon reviews \cite{mcauley2013amateurs} and Yelp reviews \cite{yelp_dataset}. In the Amazon dataset, we focus on the Musical Instruments category, analyzing text comments, user votes, and product information. In the Yelp dataset, we study Greenwood City owing to its relatively abundant number of products to the number of users. In the Amazon context, users with at least 20 votes are suitable samples; those with over 70\% useful votes \cite{kumar2018rev2} are benign, and others are spam. In Yelp, users with an average helpful score above 30\% are considered benign.

\subsubsection{Graph Construction}

We construct heterogeneous graphs for the Amazon and Yelp review datasets, consisting of two node types (user and product) and three edge types \cite{mcauley2013amateurs}. The first edge type, U$\leftrightarrow$U, connects users with high similarities, such as similar review texts or shared product purchases. The second edge type, U$\leftrightarrow$P, links users and products when a user has rated or reviewed a product. The third edge type, P$\leftrightarrow$P, connects products with similar descriptions or categories. These graphs capture various relationships and interactions, comprehensively representing the review data. Table \ref{tb_graph_stat} shows the graph statistics, with additional construction details in the supplementary material.

\begin{table}
\centering
\caption{Graph Statistics}\label{tb_graph_stat}
\begin{tabular}{
  @{}
  c
  S[table-format=4.0, group-separator={,}, group-four-digits=true]@{\,}l
  S[table-format=4.0, group-separator={,}, group-four-digits=true]
  S[table-format=7.0, group-separator={,}, group-four-digits=true]
  S[table-format=5.0, group-separator={,}, group-four-digits=true]
  S[table-format=6.0, group-separator={,}, group-four-digits=true]
  @{}} 
\toprule
\toprule
& \multicolumn{3}{c}{Node Type} & \multicolumn{3}{c}{Edge Type} \\
\cmidrule(lr){2-4} \cmidrule(lr){5-7}
{Dataset} & \multicolumn{2}{c}{User (Fraud)} & {Product} & {U$\leftrightarrow$U} & {U$\leftrightarrow$P} & {P$\leftrightarrow$P} \\
\midrule
Amazon & 7017 & {(9.7\%)} & 4684 & 535244 & 24338 & 101678 \\
Yelp & 5052 & {(19\%)} & 649 & 1131080 & 12624 & 95856 \\
\bottomrule
\bottomrule
\end{tabular}
\end{table}

\begin{table*}[ht]
\centering
\caption{Performance Improvement vs. Graph Size Increment}\label{augmented_graph_size_vs_performance}
\begin{tabular}{@{}cl|cccccc@{}} 
\toprule
\toprule
\multirow{10}{*}[9.8ex]{\rotatebox[origin=c]{90}{\parbox{3.5cm}{\centering Dataset\\(IR)}}} & \multicolumn{1}{c}{Graph Statistics} & \multicolumn{1}{|c}{$\textbf{h}_v \in \mathbb{R}^{d}$} & \multicolumn{1}{c}{\#U$\leftrightarrow$U edges} & \multicolumn{1}{c}{\#U$\leftrightarrow$P edges} & \multicolumn{1}{c}{graph size} & \multicolumn{1}{c}{AUC-PRC} & \multicolumn{1}{c}{$\dfrac{\text{AUC-RPC improvement}}{\text{graph size increment}}$}\\
\cmidrule(lr){2-8}
& Method $\diagdown$ unit & dim & amount & amount & megabyte & score & $\dfrac{\text{score}}{\text{megabyte}}$ \\
\midrule
\multirow{10}{*}[7.3ex]{\rotatebox[origin=c]{90}{\parbox{3.5cm}{\centering Amazon\\(IR=0.4)}}} & Original & 23 & 535,244 & 12,169 & 13.10  & 41.51 $\scriptstyle{\pm0.048}$ & - \\
& Oversampling & 23 & 774,738 $\scriptstyle{(+44.74\%)}$ & 13,989 $\scriptstyle{(+14.96\%)}$ & 15.51 $\scriptstyle{(+18.40\%)}$ & 36.70 $\scriptstyle{\pm0.030}$ &  -1.99 (\mbox{\larger[-1]$\searrow$}) \\
& SMOTE & 23 & 735,830 $\scriptstyle{(+37.48\%)}$ & 17,153 $\scriptstyle{(+40.96\%)}$ & 15.02 $\scriptstyle{(+14.66\%)}$ & 35.30 $\scriptstyle{\pm0.034}$ &  -3.23 (\mbox{\larger[-1]$\searrow$}) \\
& GraphSMOTE & 256 & 5,905,848 $\scriptstyle{(+1003\%)}$ & 12,169 $\scriptstyle{(-)}$ & 93.72 $\scriptstyle{(+615.4\%)}$ & 43.87 $\scriptstyle{\pm0.032}$ &  +0.03 (\mbox{\larger[-1]$\nearrow$}) \\
& \cellcolor{gray!30} \textbf{FlashGAN} & \cellcolor{gray!30} 256 & \cellcolor{gray!30} 537,778 $\scriptstyle{(+0.473\%)}$ & \cellcolor{gray!30} 15,321 $\scriptstyle{(+25.90\%)}$ & \cellcolor{gray!30} 23.3 $\scriptstyle{(+77.86\%)}$ & \cellcolor{gray!30} \textbf{52.31} $\scriptstyle{\pm0.016}$ &  \cellcolor{gray!30} \textbf{+1.06 (\mbox{\larger[-1]$\nearrow$})} \\
\midrule
\multirow{10}{*}[7.3ex]{\rotatebox[origin=c]{90}{\parbox{3.5cm}{\centering Yelp\\(IR=1.0)}}} & Original & 32 & 1,131,080 & 12,624 & 19.90  & 37.15 $\scriptstyle{\pm0.024}$ & - \\
& Oversampling & 32 & 1,542,868 $\scriptstyle{(+36.41\%)}$ & 16,580 $\scriptstyle{(+31.34\%)}$ & 26.50 $\scriptstyle{(+33.17\%)}$ & 35.81 $\scriptstyle{\pm0.014}$ & -0.203 (\mbox{\larger[-1]$\searrow$}) \\
& SMOTE & 32 & 1,608,360 $\scriptstyle{(+33.16\%)}$ & 18,453 $\scriptstyle{(+46.17\%)}$ & 27.60 $\scriptstyle{(+38.60\%)}$ & 36.93 $\scriptstyle{\pm0.030}$ & -0.029 (\mbox{\larger[-1]$\searrow$}) \\
& GraphSMOTE & 256 & 2,083,964 $\scriptstyle{(+84.25\%)}$ & 12,624 $\scriptstyle{(-)}$ & 41.30 $\scriptstyle{(+107.5\%)}$ & 37.25 $\scriptstyle{\pm0.010}$ & +0.005 (\mbox{\larger[-1]$\nearrow$}) \\
& \cellcolor{gray!30} \textbf{FlashGAN} & \cellcolor{gray!30} 256 & \cellcolor{gray!30} 1,240,340 $\scriptstyle{(+9.659\%)}$ & \cellcolor{gray!30} 18,923 $\scriptstyle{(+49.90\%)}$ & \cellcolor{gray!30} 28.70 $\scriptstyle{(+44.22\%)}$ & \cellcolor{gray!30} \textbf{40.41} $\scriptstyle{\pm0.013}$ & \cellcolor{gray!30} \textbf{+0.370 (\mbox{\larger[-1]$\nearrow$})} \\
\bottomrule
\bottomrule
\end{tabular}
\end{table*}

\subsubsection{Baselines}
To validate the proposed FlashGAN framework, we compare it with several representative approaches for addressing class imbalance problems. These include traditional oversampling methods adapted for graphs, such as Oversampling and SMOTE \cite{chawla2002smote}, classic loss adjustment techniques like Reweight and Focal Loss \cite{focal_loss}, and various GNN-based oversampling strategies. For Oversampling and SMOTE, we establish connections between the synthetic nodes and the neighboring nodes of the source nodes referenced during their generation. Among the GNN baselines, with the exception of GraphSMOTE, we retain the U$\leftrightarrow$U edge and transform the heterogeneous graph dataset into a homogeneous graph containing only user nodes, as this approach yields better experimental results compared to other methods of transforming heterogeneous graphs into homogeneous ones. The technical details of the GNN methods are described as follows:

\begin{itemize}
\item
    PC-GNN \cite{10.1145/3442381.3449989} uses a label-balanced sampler to select nodes and edges and a neighborhood sampler to pick neighbor candidates. The model aggregates information from these neighbors to derive the target node’s final representation.
    
\item
    ImGAGN \cite{Qu2021ImGAGN} introduces the GraphGenerator that simulates attributes and network structure distribution for minority class nodes. However, it is still limited to using GCN layers and only supports homogeneous graph data.
\item
    GraphSMOTE \cite{zhao2021graphsmote} extends the capabilities of SMOTE to accommodate homogeneous graphs with only one edge type. We generate spam user nodes and establish connections exclusively on the U$\leftrightarrow$U edges within the original graph.
\item
    GraphENS \cite{park2022graphens} addresses class imbalance by synthesizing ego networks for minor class nodes and integrating a saliency-based node mixing technique.

\item 
    GraphSHA \cite{li2023graphsha} addresses class imbalance by enlarging the decision boundaries of minor classes through the synthesis of harder minor samples, featuring a SemiMixup module.

\end{itemize}

\begin{figure}[h]
\centering
\subfloat[Amazon: AUC-PRC]{
\includegraphics[width=0.23\textwidth]{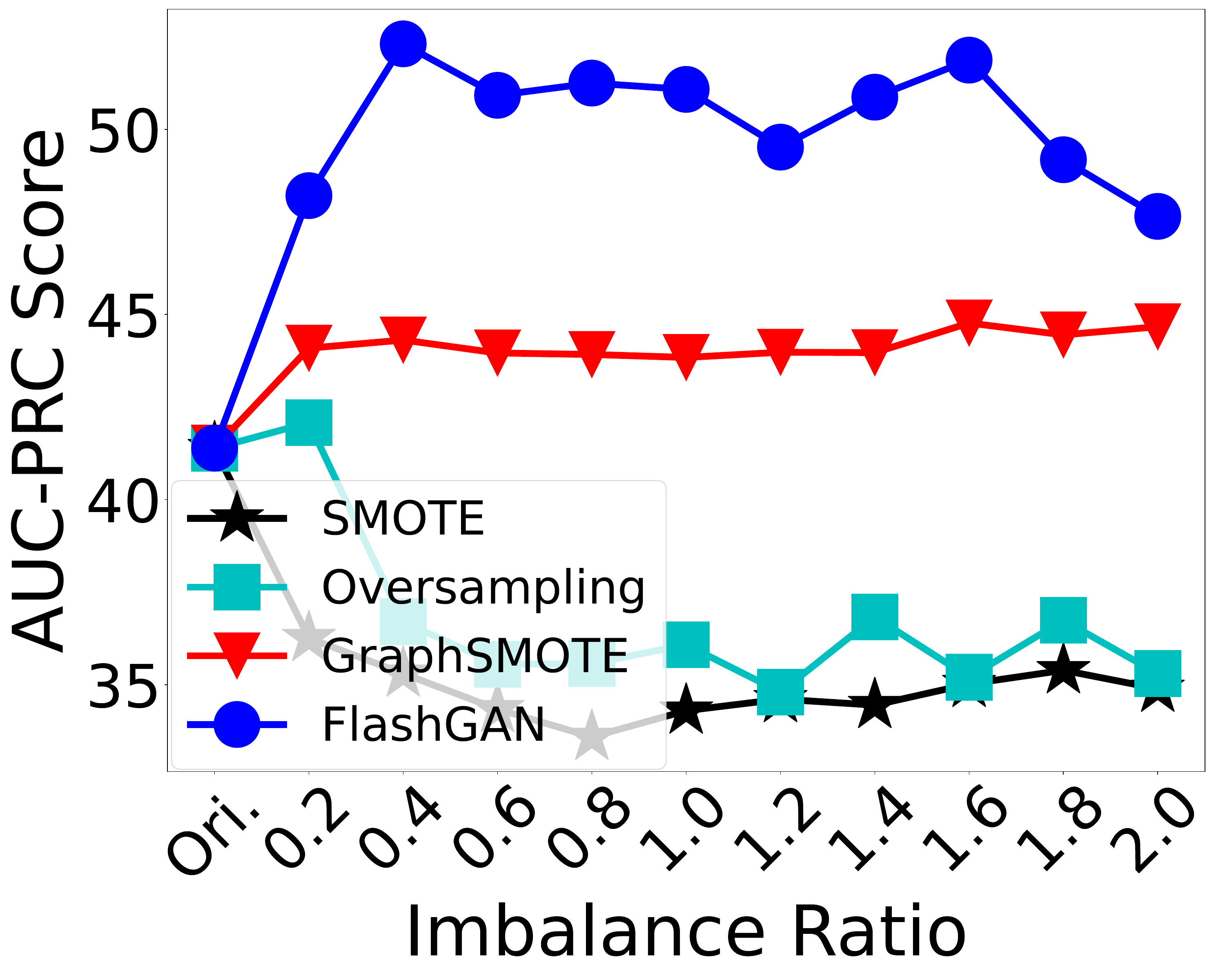}
}
\subfloat[Yelp: AUC-PRC]{
\includegraphics[width=0.23\textwidth]{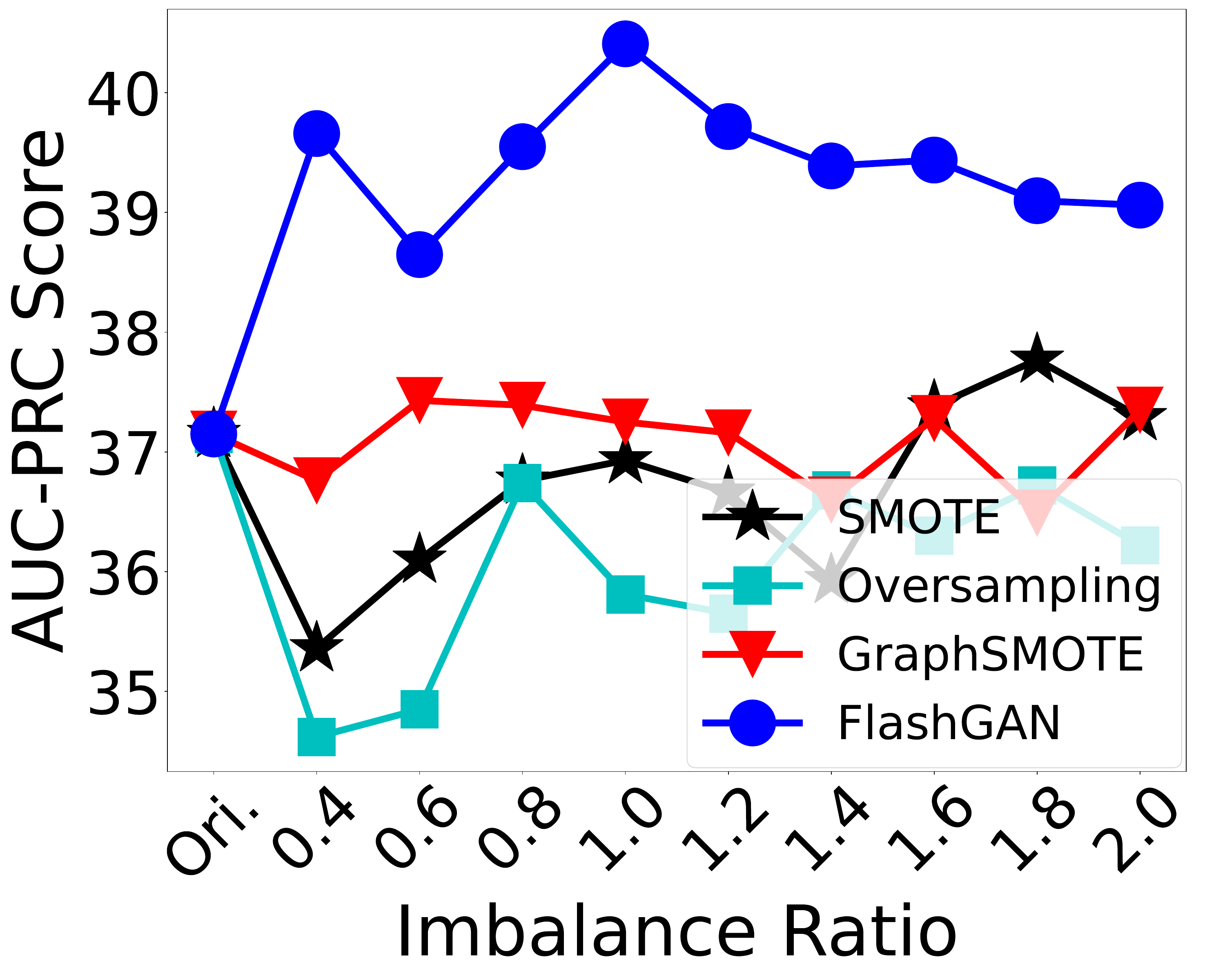}
}
\quad
\subfloat[Amazon: F-score]{
\includegraphics[width=0.23\textwidth]{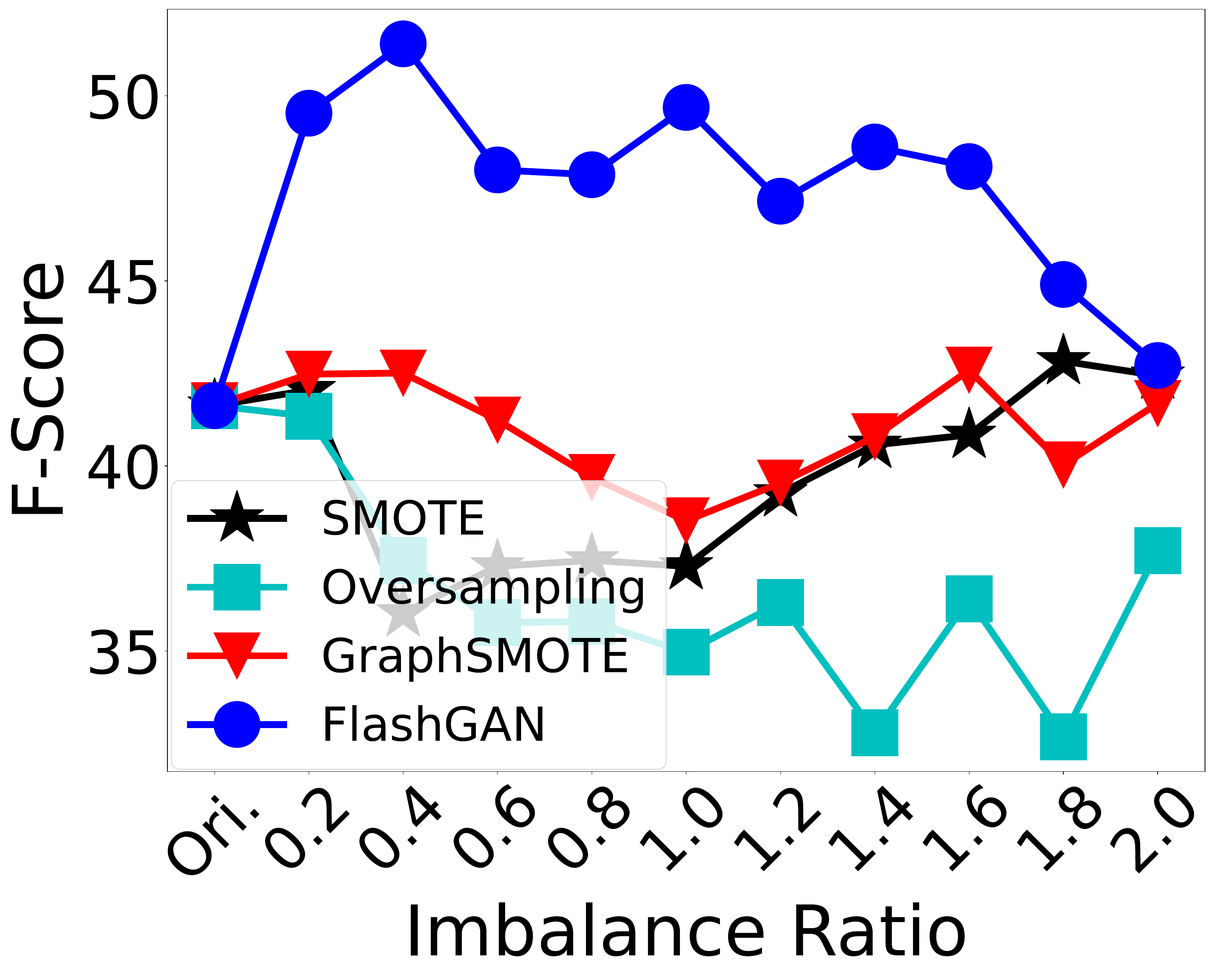}
}
\subfloat[Yelp: F-score]{
\includegraphics[width=0.23\textwidth]{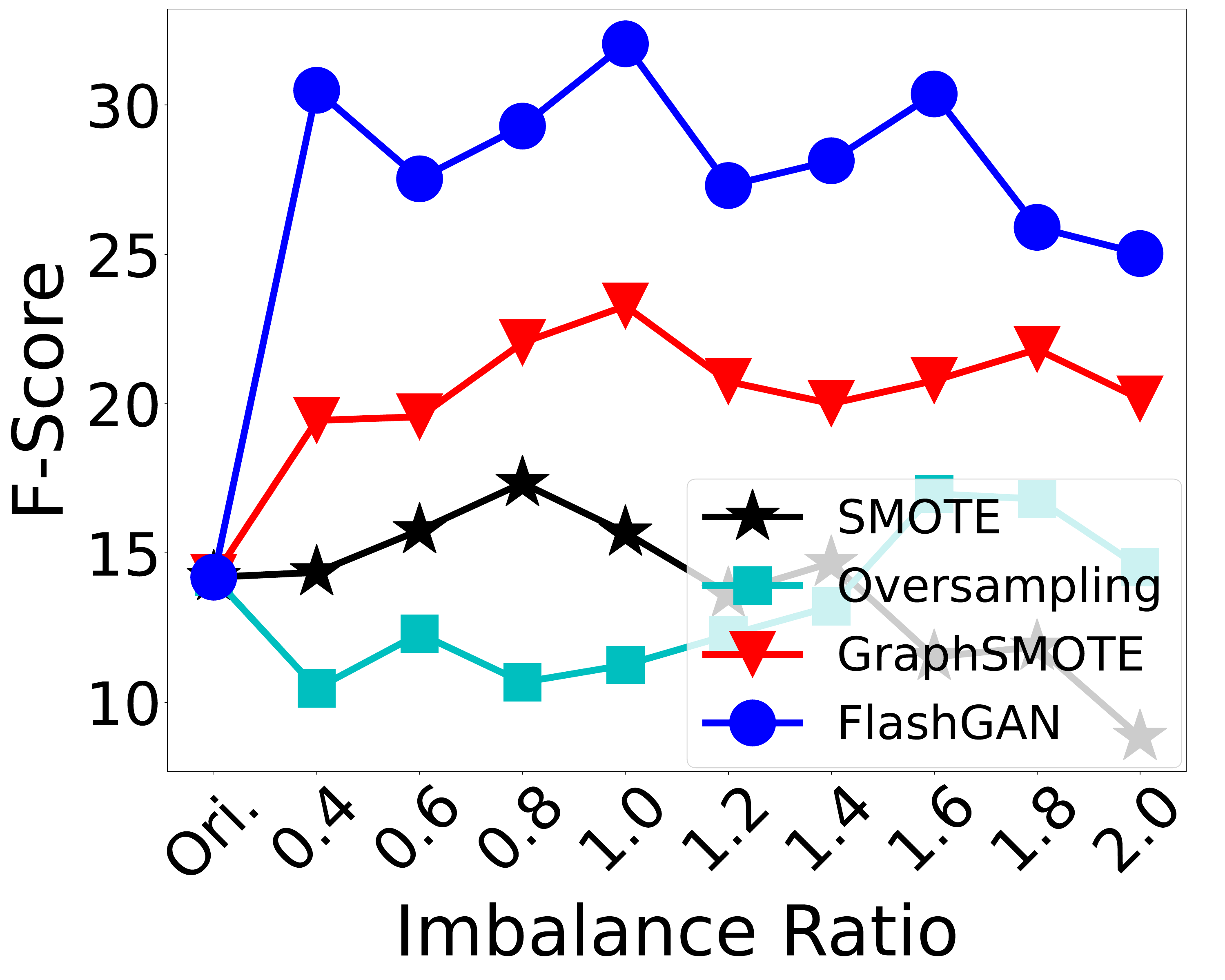}
}
\caption{Influence of Imbalance Ratio}\label{amazon_yelp_fig_up_scale}
\end{figure}

\subsubsection{Evaluation Metrics}

We evaluate our model using AUC-ROC, AUC-PRC, F-Score, and Accuracy. AUC-ROC \cite{bradley1997use} is widely used, but it can overestimate performance on imbalanced classes. In such cases, AUC-PRC \cite{davis2006relationship} is preferred for a more accurate assessment.

\subsection{Imbalanced Classification Performance}

Table \ref{tb_result} presents the experimental results of FlashGAN compared to selected baselines. We report the best performance of each baseline based on AUC-PRC metrics. On the Amazon dataset, FlashGAN achieves a 17\% improvement in AUC-PRC and a 22\% improvement in F-Score compared to the second-best method. On Yelp, it shows a 2.6\% improvement in AUC-PRC and a 29\% improvement in F-scores. These results highlight FlashGAN's superiority in key metrics for predicting minority class data, significantly enhancing the detection of minority nodes. In addition to FlashGAN’s default one-hop subgraph sampling, we also tested random walk subgraph sampling, detailed in the supplementary material, to evaluate the impact of subgraph sampling strategies on FlashGAN training. This variant, denoted as FlashGAN-R.Walk in the table, did not result in better classification performance on minority class data compared to the one-hop subgraph approach. These findings support the effectiveness of FlashGAN's local augmentation strategy and the choice of one-hop subgraph sampling.

\subsection{Influence of Imbalance Ratio}
In this subsection, we evaluate the performance of FlashGAN and other oversampling-based methods on node classification by embedding varying numbers of synthetic nodes into the graph. The line chart in Figure \ref{amazon_yelp_fig_up_scale} shows the AUC-PRC and F-score metrics, starting with the original graph without synthetic nodes (Ori.) and extending to imbalance ratios up to 2.0. We used HGT for node classification and reported average results. FlashGAN consistently excels across various imbalance ratios in both AUC-PRC and F-score. Baseline methods like Reweight and PC-GNN were excluded from this comparison since they are not oversampling techniques and are unaffected by additional data. Additionally, GraphENS, designed to generate ego networks for real minority nodes, is not suitable for comparison by imbalance ratio. ImGAGN and GraphSHA were also excluded due to the requirement that the imbalance ratio be greater than 1 for their implementation.

\begin{figure}[h]
\centering
\subfloat[Training Losses of FlashGAN]{
\includegraphics[width=0.24\textwidth]{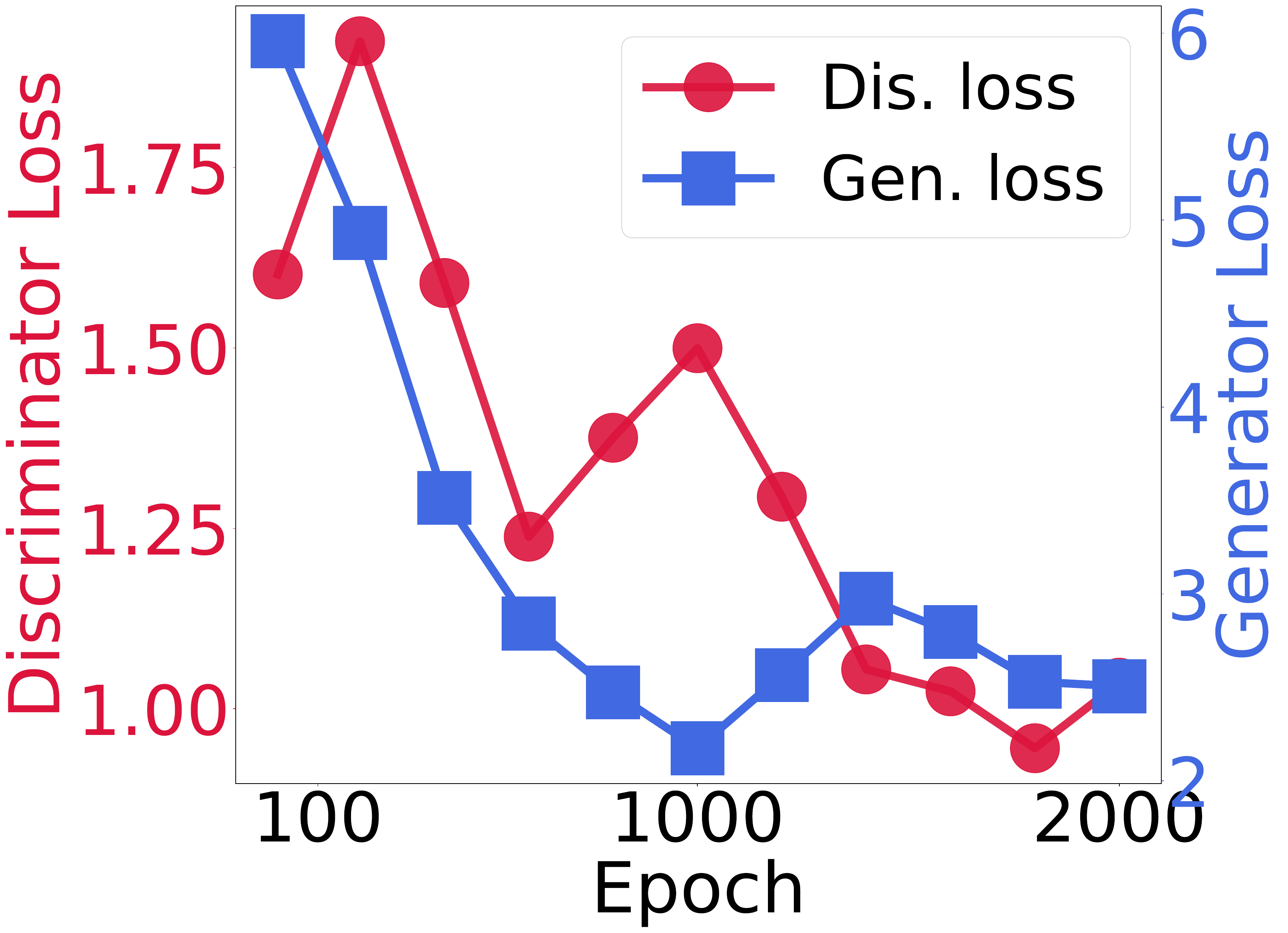}
}
\subfloat[Effectiveness of the Edge Filter]{
\includegraphics[width=0.23\textwidth]{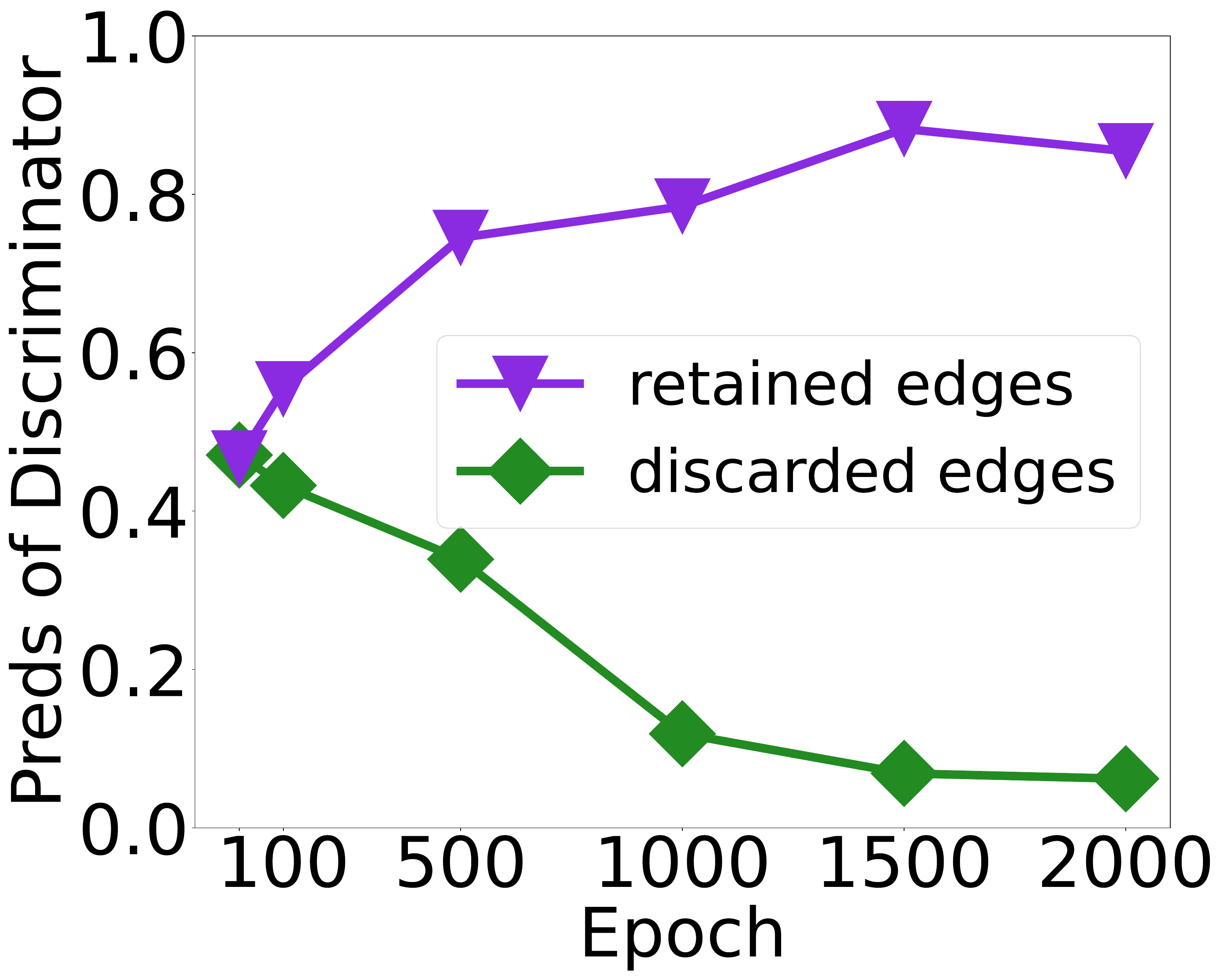}
}
\caption{Investigation of the Synthetic Edge Filter}\label{chars_of_syn_edges}
\end{figure}

\subsection{Investigation of the Synthetic Edge Filter}
We investigate the impact of the synthetic edge filter $\Delta_r$ on improving minority node classification performance in downstream tasks. Specifically, given a synthetic edge filter $\Delta_r$ designed for type $r$, which initially selects $n$ synthetic edges based on a threshold $\eta_r$, we also assess alternative edge selection strategies. Rather than relying on the threshold $\eta_r$, we consider two alternatives: randomly selecting $n$ synthetic edges and choosing the $n$ synthetic edges with the lowest probabilities from the potential set. This methodology allows us to assess the filter's effectiveness, with results presented in Table \ref{tb_result} for FlashGAN-LPnE (least probable $n$ edges) and FlashGAN-RSSE (randomly sampled synthetic edges). The findings indicate that edges with the highest probabilities markedly improve classification performance and confirm the filter's role in selecting edges most akin to real edges. Figure 3 illustrates the experimental results of training FlashGAN on the Amazon dataset. Figure 3(a) displays the loss curves for the generator and discriminator as the training progress. Figure 3(b) demonstrates the synthetic edge filter's ability to learn to differentiate which potential edges can be retained in the subgraph. It shows how these retained and discarded potential edges, on average, are classified by the discriminator as real (1) or fake (0), respectively.

\subsection{Effectiveness measure of the synthetic data}

Since FlashGAN leverages both node and edge information for co-training, it effectively models the edge behavior of minority nodes. Table \ref{augmented_graph_size_vs_performance} compares FlashGAN with baseline methods that add nodes and edges to the Amazon and Yelp datasets, evaluating the impact of augmented graph size on performance. FlashGAN outperforms other oversampling methods, achieving the highest AUC-PRC scores with an Imbalance Ratio of 0.4 for Amazon and 1.0 for Yelp. Additionally, Table \ref{augmented_graph_size_vs_performance} shows that FlashGAN requires minimal synthetic data to significantly improve AUC-PRC scores, highlighting the efficiency of its data augmentation in enhancing minority class classification.

\section{Conclusions}
We propose FlashGAN, a novel topology-aware framework designed to generate synthetic nodes and improve predictions for minority classes in graph data. FlashGAN employs subgraph sampling and uses subgraphs as the fundamental training units, ensuring that the framework can effectively handle graph datasets of varying sizes. By conducting node augmentation within local subgraphs, FlashGAN avoids the need to sample from the highly imbalanced positive and negative edges across the entire graph dataset, thereby mitigating the edge imbalance issue. Additionally, the integration of a synthetic edge filter into the conventional GAN framework enhances its capability by directly considering both node attributes and edge information during the adversarial learning process. This approach enables the model to optimally incorporate synthetic nodes into subgraphs, significantly improving downstream node classification tasks. Moreover, the synthetic edge filter eliminates the need for post-experimental tuning of edge threshold hyperparameters, which was previously required to effectively integrate isolated synthetic nodes into the original graph. Experimental results demonstrate FlashGAN's superiority over baseline models in key metrics for minority class prediction, achieving these improved outcomes with minimal additional data integration compared to other methods.

\bibliographystyle{ACM-Reference-Format}

\bibliography{bibfile}


\appendix

\section{Pseudocode}

The training pipeline of FlashGAN is outlined in Algorithm $\ref{FlashGAN_algorithm}$. To prevent data leakage, labels of testing nodes are masked throughout the training of FlashGAN. After training, the FlashGAN generator produces synthetic nodes and edges, which are incorporated into the training set for downstream node classification, enhancing the classification performance on minor nodes in the testing data. The training process of FlashGAN begins by extracting preliminary node representations via a feature extractor. Sampled subgraphs are then augmented with synthetic nodes and edges, as detailed in Algorithm 2. These augmented subgraphs are then fed into the generator and discriminator, where parameters are iteratively updated using their respective loss functions.

\begin{algorithm}
\caption{FlashGAN}\label{FlashGAN_algorithm}
\SetKwInput{KwRequire}{Require}
\KwRequire{Graph data $G = (V, E, \textbf{X}_V)$}
Extract graph feature $G (V, E, \textbf{H}_V) \xleftarrow[~]{\Theta} G(V, E, \textbf{X}_V)$ by \eqref{eq_feature_extract}\; 
Initialize discriminator $\textsc{Dis}$ and generator $\textsc{Gen}(\Omega, \Delta_r)$\;
\While{not converge}{
    \For{number of epochs}{
        \tcp{Train generator}
        Sample $m$ filtered augmented subgraphs $SG_{\Delta}$ from $G$ through the Subgraph Augmentation $\beta$ procedure $\{SG_{\Delta}^{(1)}, ..., SG_{\Delta}^{(m)}\} \xleftarrow[~]{\beta} (G,\textsc{Gen}, \textsc{Dis})$\;
        Compute $\mathcal{L}_{\textsc{Gen}}^{(i)}$ of each $SG_{\Delta}^{(i)}$ by \eqref{eq_generator_loss_function}\;
        Update the generator $\textsc{Gen}$ with $\nabla \sum\limits_{i=1\vphantom{r}}^{m\vphantom{\mathcal{R}}} \mathcal{L}_{\textsc{Gen}}^{(i)}$\;
        \tcp{Train discriminator}
        Compute $\mathcal{L}_{\textsc{Dis}}^{(i)}$ of each $SG_{\Delta}^{(i)}$ by (14)\;
        Update the discriminator $\textsc{Dis}$ with $\nabla \sum\limits_{i=1\vphantom{r}}^{m\vphantom{\mathcal{R}}} \mathcal{L}_{\textsc{Dis}}^{(i)}$\;
    }
}
\end{algorithm}

\begin{algorithm}
\caption{Subgraph Augmentation $\beta$}\label{Subgraph_augmentation}
\SetKwInput{KwRequire}{Require}
\SetKwInput{KwInput}{Input}
\SetKwInput{KwOutput}{Output}

\KwRequire{Batchsize $m$ for the filtered augmented subgraphs, generator $\textsc{Gen}$, and discriminator $\textsc{Dis}$}
\KwInput{Graph $G$}
\KwOutput{A batch of filtered augmented subgraphs, $\mathcal{S}$}
$\{SG^{(1)}, ..., SG^{(m)}\} \xleftarrow[~]{} G$\;
\tcp{Sample $m$ induced subgraphs $SG$ from $G$}
\textbf{Parallel }\For{$SG^{(i)}$}{

    $SG_{\text{aug}}^{(i)} \xleftarrow[~]{\Omega} SG^{(i)}$ by \eqref{eq_augmented_subgraph}\; 
    \tcp{Insert synthetic nodes of minority class}
    $\eta_r^{*(i)} \xleftarrow[~]{B_r, \Delta_r} F_{\text{real}}^{(i)}$ by \eqref{optimized_eta} and \eqref{binary_step_function}\;
    \tcp{Get ths for synthetic edges filtering}
    $SG_{\Delta}^{(i)} \xleftarrow[~]{\Delta_r} (SG_{\text{aug}}^{(i)}, F_{\text{potential}}^{(i)}, \eta_r^{*(i)})$ by \eqref{eq_syn_edge_filter}\; 
    \tcp{Filter synthetic edges}
}
$\mathcal{S} \xleftarrow[~]{} \{SG_{\Delta}^{(1)}, ..., SG_{\Delta}^{(m)}\}$\;
\tcp{Return the filtered augmented subgraph batch}
\end{algorithm}

\clearpage

\section{Details of Graph Datasets}
\subsection{Datasets}
We employ FlashGAN to address class imbalance in the Amazon and Yelp review datasets. In the Amazon dataset, focused on the Musical Instruments category, we analyze text comments, user votes, and product information, classifying users with over 70\% useful votes as benign and others as spam. For the Yelp dataset, centered on Greenwood City, users with an average helpful score above 30\% are considered benign. Tables \ref{Amazon_User_features}, \ref{Amazon_Product_features}, \ref{Yelp_user_features}, and \ref{Yelp_product_features} detail the features extracted for user and product nodes in both datasets, which were derived from raw data to characterize nodes and construct edges. To avoid information leakage, features related to user labeling, such as spam and helpful votes, were removed and are marked in italics in the tables.

\subsection{Graph Construction}
We construct a heterogeneous graph for the Amazon review dataset and the Yelp review dataset, respectively, which consists of two node types (user and product) and three edge types \cite{mcauley2013amateurs}. The first edge type, U$\leftrightarrow$U, connects users with high similarities, such as those with similar review texts or who have purchased (reviewed) the same product. For the Amazon dataset, we connect two users if their review similarities measured by TF-IDF exceed 0.6; in the Yelp dataset, users are connected with others whose reviews are among the top 1\% in similarity. The second edge type, U$\leftrightarrow$P, connects users and products if a user has rated or reviewed a product. The third edge type, P$\leftrightarrow$P, connects products that are similar in their product description or belong to the same category. These heterogeneous graphs capture various relationships and interactions among users and products, comprehensively representing these review data.

\section{Implementation Details}

\subsection{Configurations}
To minimize the impact of outliers on the experimental results, we conduct experiments using ten random seeds, ranging from 0 to 9, for each configuration. The average of these results serves as the representative value for each experimental group's outcome. Our experiments were conducted in an environment with Ubuntu 20.04.5 LTS, CUDA 11.6.2 \cite{cuda}, torch 2.0.0 \cite{PyTorch}, and dgl 1.0.1+cu117\cite{wang2019dgl}. The machine is equipped with NVIDIA RTX A6000 GPU.

\subsection{Details of FlashGAN Training}
During the Feature Extraction stage \eqref{eq_feature_extract}, nodes in the graph are divided into training, validation, and test sets in a 7:2:1 ratio. We employ the Heterogeneous Graph Transformer (HGT) to learn new node representations with a dimension of 256, enhancing the original node dimensions of 23 and 32 for the Amazon and Yelp graphs, respectively. Maxpooling is used for graph pooling, while the DCGAN architecture is kept consistent with the original study, with only the final output modified to a dimension of 256. During the synthetic node embedding generation phase, we use different settings for the number of synthetic nodes appended to each subgraph in the Amazon and Yelp datasets. In the Amazon dataset, we add 5 synthetic nodes per subgraph, whereas in the Yelp dataset, we add 3, due to the higher subgraph density in Yelp. The synthetic edge filter and the discriminator's edge classifier use a weight matrix of 256x256. Before predicting real and synthetic edges, the discriminator's edge classifier, equipped with a single-layer HGT and a head count of 1, learns new representations for nodes in the filtered augmented subgraph.

\subsection{Details of Baseline Implementation}
For the loss-adjusting approaches, we replaced HGT's original loss function with reweight and Focal Loss for training, setting the weights in reweight by the reciprocal of a class's size multiplied by a fixed factor, $k = 2$, and for Focal Loss, setting $\alpha$ to $0.25$ and $\gamma$ to $2.0$. In the implementations of PC-GNN and ImGAGN, we conducted grid searches across several hyperparameters. For PC-GNN, we varied hidden dimensions within $\{64, 128\}$, learning rates from $\{5e-3, 5e-2\}$, and training epochs from $\{10, 20, 50\}$. ImGAGN's grid search included training epochs within $\{100, 200\}$, hidden dimensions $\{128, 256\}$, generator's learning epochs $\{10, 20\}$, and learning rates $\{1e-6, 1e-5, 1e-4\}$.

For GraphENS and GraphSHA, we fixed the number of layers at 2 and set imb-ratio at 10 , using SAGE as the GNN backbone for its superior performance on Amazon and Yelp datasets. GraphSHA's grid search also included training epochs within $\{50, 100, 200\}$, learning rates $\{1e-3, 5e-3, 1e-2\}$, and utilized 'ppr' (personalized PageRank) for weighted graph generation. The imbalance ratio was set to a default of 100 at the warm-up stage.

For oversampling methods like Oversampling, SMOTE, \\ GraphSMOTE, and FlashGAN, and its variants (Random Walk, LPnE, RSSE), we utilized HGT on the constructed heterogeneous graphs, testing learning rates within $\{1e-4, 1e-3, 1e-2\}$ and Adam scheduler’s pct-start within $\{0.3, 0.5, 0.7\}$, maintaining other HGT hyperparameters at a hidden dimension of $256$, two layers, and heads set to $4$ for Amazon and $1$ for Yelp.

\subsection{Random Walk Subgraph Sampling}
We explore the use of random walk subgraph sampling as an alternative to FlashGAN's default one-hop subgraph sampling method. Our goal is to determine whether training FlashGAN with subgraphs sampled via random walks, as opposed to one-hop subgraphs, better facilitates the generation of synthetic nodes that improve minor node classification in downstream tasks. With this motivation, we aim to compare the impact on FlashGAN’s training when the sampled subgraphs contain a similar number of nodes but differ in their spatial distribution within the graph. To this end, we calculated the average size of one-hop subgraphs centered on each user node in the Amazon and Yelp datasets. We found that there are 2,500 valid subgraphs in the Amazon dataset and 4,584 in the Yelp dataset (each containing both user and product nodes), with average sizes of 88 and 253 user nodes, respectively. To ensure that the random walk subgraph sampling yields subgraphs of comparable size to these one-hop subgraphs, we adjusted the corresponding step length to match the average subgraph size between the two methods.

\clearpage

\begin{table*}
    \centering
    \caption{User Features of Amazon Graph Dataset}
    \label{Amazon_User_features}
    \begin{tabular}{l||l}
        \toprule
        \toprule
        \multicolumn{2}{l}{\textbf{User Features}} \\
        \cmidrule{1-2}
        Feature name & Feature meaning \\
        \midrule
        user ID & Index of the user \\
        $\emph{spam}$ & 1 if the user was labeled as spam user; else 0 \\
        product\_cnt & Number of products rated by the user\\
        1-star$\sim$5-star & Number of each rating level given by the user \\
        min/max & Min and max helpful score of user review \\
        mean/median & Mean and median of user review helpful score \\
        high/low & Highest and lowest helpful score of user review \\
        review\_cnt & Number of review \\
        rate\_entropy & Entropy of the rating level distribution given by the user \cite{https://doi.org/10.48550/arxiv.2005.10150}\\
        time\_dif & Difference between the earliest and the latest review date \cite{Fei_Mukherjee_Liu_Hsu_Castellanos_Ghosh_2021, Mukherjee_Venkataraman_Liu_Glance_2021, 10.1145/2783258.2783370} \\
        same\_day & 1 if the first and the latest review is on the same day; else 0 \\
        time\_entropy & Entropy of the review time gap distribution of the user \cite{https://doi.org/10.48550/arxiv.2005.10150} \\
        word\_cnt & Avg words of review \\
        name\_len & Length of username \\
        $\emph{(un)help\_cnt}$ & Total number of helpful and unhelpful votes evaluated by other users. \\
        $\emph{(un)help\_ratio}$ & Ratio of helpful and unhelpful votes evaluated by other users. \\
        $\emph{(un)help\_mean}$ & Mean of helpful and unhelpful votes evaluated by other users. \\
        POS & Ratio of positive words \cite{10.5555/2283696.2283811} \\
        NEG & Ratio of negative words \cite{10.5555/2283696.2283811} \\
        NEU & Ratio of neutral words \cite{10.5555/2283696.2283811} \\
        \bottomrule
    \end{tabular}

\end{table*}

\begin{table*}
    \centering
    \caption{Product Features of Amazon Graph Dataset}
    \label{Amazon_Product_features}
    \begin{tabular}{l||l}
        \toprule
        \toprule
        \multicolumn{2}{l}{\textbf{Product Features}} \\
        \cmidrule{1-2}
        Feature name & Feature meaning \\
        \midrule
        product ID & Index of the product \\
        user\_cnt & Number of users that reviewed the product \\
        1-star$\sim$5-star & Number of each rating level given by users \\
        min/max & Min and max score of user ratings \\
        mean/median & Mean and median of user rating scores \\
        high/low & Highest and lowest score of user ratings \\
        review\_cnt & Total number of the product's reviews \\
        rate\_entropy & Entropy of the rating level distribution reviewed by users \cite{https://doi.org/10.48550/arxiv.2005.10150}\\
        time\_dif & Difference between the earliest and the latest review date  \cite{Fei_Mukherjee_Liu_Hsu_Castellanos_Ghosh_2021, Mukherjee_Venkataraman_Liu_Glance_2021, 10.1145/2783258.2783370} \\
        same\_day & 1 if the first and latest product reviews are on the same day; otherwise, 0. \\
        time\_entropy & Entropy of the time gap distribution of user reviews \cite{https://doi.org/10.48550/arxiv.2005.10150}\\
        word\_cnt & Avg words of review \\
        product\_kurtosis & Kurtosis of product sales\\
        product\_price & Price of product\\
        TF-IDF & TF-IDF vector or product\\
        \bottomrule
        \bottomrule
    \end{tabular}

\end{table*}

\begin{table*}
\centering
    \caption{User Features of Yelp Graph Dataset}
    \label{Yelp_user_features}
    \begin{tabular}{l||l||l}
        \toprule
        \toprule
        \multicolumn{3}{l}{\textbf{User Features}} \\
        \cmidrule{1-3}
        Name & Statistics & Meaning\\
        \midrule
        \multirow{2}{*}{user ID} & \multirow{2}{*}{-} & \multirow{2}{*}{Index of the user} \\ & & \\
        \multirow{2}{*}{review\_cnt} & \multirow{2}{*}{-} & \multirow{2}{*}{Number of user's reviews} \\ & & \\
        \multirow{2}{*}{$\emph{useful}$} & \multirow{2}{*}{-} & \multirow{2}{*}{Number of useful votes to the user's reviews given by other users} \\ & & \\
        \multirow{2}{*}{funny} & \multirow{2}{*}{-} & \multirow{2}{*}{Number of funny votes to the user's reviews given by other users} \\ & & \\
        \multirow{2}{*}{cool} & \multirow{2}{*}{-} & \multirow{2}{*}{Number of cool votes to the user's reviews given by other users} \\ & & \\
        
        \multirow{2}{*}{star} & \multirow{2}{*}{Avg} & \multirow{2}{*}{Star scores given by the user} \\ & & \\
        \multirow{2}{*}{MNR} & \multirow{2}{*}{-} & \multirow{2}{*}{Max number of user's reviews in a day \cite{10.1145/2487575.2487580}} \\ & & \\
        \multirow{2}{*}{PR} & \multirow{2}{*}{-} & \multirow{2}{*}{Ratio of user's positive reviews (4-5 stars) \cite{10.1145/2487575.2487580}} \\ & & \\
        \multirow{2}{*}{NR} & \multirow{2}{*}{-} & \multirow{2}{*}{Ratio of user's negative reviews (1-2 stars) \cite{10.1145/2487575.2487580}} \\ & & \\
        \multirow{2}{*}{RD} & \multirow{2}{*}{Weighted Sum, Sum, Avg} & \multirow{2}{*}{Rating deviations across the user's reviews \cite{10.1145/2487575.2487580}} \\ & & \\ 
        \multirow{2}{*}{BST} & \multirow{2}{*}{-} & 1 if the first and the latest reviews that the product received is on the same day; \\ & & otherwise 0 \cite{10.1145/2783258.2783370}\\
        
        \multirow{2}{*}{ERD} & \multirow{2}{*}{-} & \multirow{2}{*}{Entropy rating distribution of the user's reviews \cite{10.1145/2487575.2487580}} \\ & & \\         
        \multirow{2}{*}{ETG} & \multirow{2}{*}{-} & \multirow{2}{*}{Entropy of temporal gaps of the user's reviews \cite{10.1145/2487575.2487580}} \\ & & \\

        \multirow{2}{*}{RCS} & \multirow{2}{*}{Maximum, Avg} & \multirow{2}{*}{Cosine similarities across the user's reviews} \\ & & \\

        \multirow{2}{*}{PCW} & \multirow{2}{*}{Sum, Avg} & \multirow{2}{*}{Percentage of ALL-capital words across the user's reviews \cite{10.5555/2283696.2283811}} \\ & & \\
        \multirow{2}{*}{PC} & \multirow{2}{*}{Sum, Avg} & \multirow{2}{*}{Percentage of capital letters across the user's reviews \cite{10.5555/2283696.2283811}} \\ & & \\
        \multirow{2}{*}{PP1} & \multirow{2}{*}{Sum, Avg} & \multirow{2}{*}{Percentage of 1st-personal pronouns across the user's reviews \cite{10.5555/2283696.2283811}} \\ & & \\
        \multirow{2}{*}{RES} & \multirow{2}{*}{Sum, Avg} & \multirow{2}{*}{Percentage of exclamation sentences (!) across the user's reviews \cite{10.5555/2283696.2283811}} \\ & & \\
        \multirow{2}{*}{L} & \multirow{2}{*}{Sum, Avg} & \multirow{2}{*}{User's reviews length \cite{10.5555/2283696.2283811}} \\ & & \\

        \multirow{2}{*}{PW} & \multirow{2}{*}{Sum, Avg} & \multirow{2}{*}{Ratio of the positive words across the user's reviews \cite{10.5555/2283696.2283811}} \\ & & \\
        \multirow{2}{*}{NW} & \multirow{2}{*}{Sum, Avg} & \multirow{2}{*}{Ratio of the negative words across the user's reviews \cite{10.5555/2283696.2283811}} \\ & & \\
        \multirow{2}{*}{OW} & \multirow{2}{*}{Sum, Avg} & \multirow{2}{*}{Ratio of the objective words across the user's reviews \cite{10.5555/2283696.2283811}} \\ & & \\
        \bottomrule
        \bottomrule
    \end{tabular}
\end{table*}

\begin{table*}
\centering
    \caption{Product Features of Yelp Graph Dataset}
    \label{Yelp_product_features}
    \begin{tabular}{l||l||l}
        \toprule
        \toprule
        \multicolumn{3}{l}{\textbf{Product Features}} \\
        \cmidrule{1-3}
        Name & Statistics & Meaning\\
        \midrule
        \multirow{2}{*}{product ID} & \multirow{2}{*}{-} & \multirow{2}{*}{Index of the product} \\ & & \\
        \multirow{2}{*}{review\_cnt} & \multirow{2}{*}{-} & \multirow{2}{*}{Total number of the product's reviews} \\ & & \\
        \multirow{2}{*}{is\_open} & \multirow{2}{*}{-} & \multirow{2}{*}{1 if the product (business) is currently open; otherwise 0} \\ & & \\
        
        \multirow{2}{*}{star} & \multirow{2}{*}{Sum, Avg, Std} & \multirow{2}{*}{Star scores that the product received from the users} \\ & & \\

        \multirow{2}{*}{MNR} & \multirow{2}{*}{-} & \multirow{2}{*}{Max number of reviews that the product received from the users in a day \cite{10.1145/2487575.2487580}} \\ & & \\
        \multirow{2}{*}{PR} & \multirow{2}{*}{-} & \multirow{2}{*}{Ratio of product's positive reviews (4-5 stars) \cite{10.1145/2487575.2487580}} \\ & & \\
        \multirow{2}{*}{NR} & \multirow{2}{*}{-} & \multirow{2}{*}{Ratio of product's negative reviews (1-2 stars) \cite{10.1145/2487575.2487580}} \\ & & \\
        
        \multirow{2}{*}{RD} & \multirow{2}{*}{Weighted Sum, Sum, Avg} & \multirow{2}{*}{Rating deviations across the product's reviews \cite{10.1145/2487575.2487580}} \\ & & \\ 
        \multirow{2}{*}{BST} & \multirow{2}{*}{-} & 1 if the first and the latest reviews that the product received is on the same day; \\ & & otherwise 0 \cite{10.1145/2783258.2783370}\\
        
        \multirow{2}{*}{ERD} & \multirow{2}{*}{-} & \multirow{2}{*}{Entropy rating distribution of the product's reviews \cite{10.1145/2487575.2487580}} \\ & & \\         
        \multirow{2}{*}{ETG} & \multirow{2}{*}{-} & \multirow{2}{*}{Entropy of temporal gaps of the product's reviews \cite{10.1145/2487575.2487580}} \\ & & \\

        \multirow{2}{*}{RCS} & \multirow{2}{*}{Maximum, Avg} & \multirow{2}{*}{Cosine similarities across the product's reviews} \\ & & \\

        \multirow{2}{*}{PCW} & \multirow{2}{*}{Sum, Avg} & \multirow{2}{*}{Percentage of ALL-capital words across the product's reviews \cite{10.5555/2283696.2283811}} \\ & & \\
        \multirow{2}{*}{PC} & \multirow{2}{*}{Sum, Avg} & \multirow{2}{*}{Percentage of capital letters across the product's reviews \cite{10.5555/2283696.2283811}} \\ & & \\
        \multirow{2}{*}{PP1} & \multirow{2}{*}{Sum, Avg} & \multirow{2}{*}{Percentage of 1st-personal pronouns across the product's reviews \cite{10.5555/2283696.2283811}} \\ & & \\
        \multirow{2}{*}{RES} & \multirow{2}{*}{Sum, Avg} & \multirow{2}{*}{Percentage of exclamation sentences (!) across the product's reviews \cite{10.5555/2283696.2283811}} \\ & & \\
        
        \multirow{2}{*}{L} & \multirow{2}{*}{Sum, Avg} & \multirow{2}{*}{Product's reviews length \cite{10.5555/2283696.2283811}} \\ & & \\

        \multirow{2}{*}{PW} & \multirow{2}{*}{Sum, Avg} & \multirow{2}{*}{Ratio of the positive words across the product's reviews \cite{10.5555/2283696.2283811}} \\ & & \\
        \multirow{2}{*}{NW} & \multirow{2}{*}{Sum, Avg} & \multirow{2}{*}{Ratio of the negative words across the product's reviews \cite{10.5555/2283696.2283811}} \\ & & \\
        \multirow{2}{*}{OW} & \multirow{2}{*}{Sum, Avg} & \multirow{2}{*}{Ratio of the objective words across the product's reviews \cite{10.5555/2283696.2283811}} \\ & & \\
        \bottomrule
        \bottomrule
    \end{tabular}
\end{table*}

\end{document}